\documentclass[letterpaper, 10 pt, journal]{IEEEtran}
\usepackage{times}

\usepackage[table]{xcolor}

\usepackage{amsmath}
\usepackage{amssymb}
\usepackage{mathtools} 
\usepackage[table]{xcolor}
\usepackage{booktabs}
\usepackage{multirow}
\usepackage{tcolorbox}

\usepackage{algorithm}
\usepackage{algpseudocode}
\usepackage{aliascnt}
\usepackage{booktabs}
\usepackage[tableposition=top]{caption}
\usepackage{enumitem}
\usepackage{float}
\usepackage{graphicx}
\usepackage{makecell}
\usepackage{multicol}
\usepackage{multirow}
\usepackage[numbers,sort&compress]{natbib}
\usepackage[all]{nowidow}
\usepackage{pgfplots}
\pgfplotsset{compat=1.18}
\usepackage{subcaption}
\usepackage{svg}
\usepackage{tabularx}
\usepackage{tikz}
\usepackage{verbatim}
\usepackage{wrapfig}

\usepackage{hyperref}
\usepackage[nameinlink, capitalize]{cleveref}

\crefname{subsection}{Sec.}{Secs.}

\newcommand{\appendixsection}[2]{%
  \newaliascnt{appsec}{section}%
  \section{#1}\label{#2}%
  \aliascntresetthe{appsec}%
  \expandafter\def\csname appsecautorefname\endcsname{Appendix}%
}

\definecolor{lightblue}{rgb}{0.93,0.95,1.0}
\definecolor{ForestGreen}{rgb}{0.13,0.55,0.13}

\definecolor{uoftblue}{RGB}{30, 55, 101} 

\definecolor{uoftsecondaryblue}{RGB}{0,127,163} 

\definecolor{uoftpurple}{RGB}{109,36,122} 
\definecolor{uoftwarmred}{RGB}{220,70,51} 
\definecolor{uoftcoolblue}{RGB}{111,199,234} 
\definecolor{uoftteal}{RGB}{0,161,137} 
\definecolor{uoftfuchsia}{RGB}{171,19,104} 
\definecolor{uoftdarkgreen}{RGB}{13,83,77} 
\definecolor{uoftyellow}{RGB}{241,197,0} 
\definecolor{uoftlightgreen}{RGB}{141,191,46} 

\definecolor{uoftcoolgray}{RGB}{208,209,201} 

\hypersetup{
    bookmarks=true,
    colorlinks=true,   
    allcolors=uoftsecondaryblue 
}

\renewcommand{\arraystretch}{1.3}
\usepackage[symbol]{footmisc}

\usepackage{amsmath,amsfonts,bm}

\def\eqref#1{equation~\ref{#1}}

\def\1{\bm{1}}

\DeclareMathAlphabet{\mathsfit}{\encodingdefault}{\sfdefault}{m}{sl}
\SetMathAlphabet{\mathsfit}{bold}{\encodingdefault}{\sfdefault}{bx}{n}

\DeclareMathOperator*{\argmax}{arg\,max}

\let\originalleft\left
\let\originalright\right
\def\left#1{\mathopen{}\originalleft#1}
\def\right#1{\originalright#1\mathclose{}}

\definecolor{OIblue}{RGB}{0,114,178}
\definecolor{OIorange}{RGB}{230,159,0}
\definecolor{OIgreen}{RGB}{0,158,115}
\definecolor{OIred}{RGB}{213,94,0}
\definecolor{OIpurple}{RGB}{204,121,167}


\begin{document}
\bstctlcite{IEEEexample:BSTcontrol}

\title{Update-Free On-Policy Steering via Verifiers}

\author{Maria Attarian$^{1, 2}$, 
        Ian Vyse$^{3}$, 
        Jasper Gerigk$^{1}$, 
        Evgenii Opryshko$^{1}$, 
        Yifan Ruan$^{1}$,
        Anas Almasri$^{1}$, 
        Sumeet Singh$^{2}$, 
        Yilun Du$^{4}$, 
        Igor Gilitschenski $^{1}$,
        Claas Voelcker$^{5}$%
\thanks{$^{1}$ University of Toronto; $^{2}$Google DeepMind; $^{3}$University of Alberta; $^{4}$Harvard University; $^{5}$University of Texas at Austin}
\thanks{Corresponding author: Maria Attarian.}%

}

\newcommand{\algoname}{{UF-OPS}}
\newcommand{\claas}[1]{{\color{blue} [\textbf{CV:}] #1}}

\maketitle

\begin{abstract} 
In recent years, Behavior Cloning (BC) has become one of the most prevalent methods for learning manipulation from human demonstrations. Despite their successes, BC policies are often brittle and struggle with precise manipulation or choosing among different demonstrator policies.
To overcome these issues, we propose \algoname, an Update-Free On-Policy Steering method that enables the robot to predict the success likelihood of its actions and adapt its strategy at execution time.
We accomplish this by training verifier functions using policy rollout data obtained during an initial evaluation of the policy.
These verifiers are subsequently used to steer the base policy toward actions with a higher likelihood of success.
Our method improves the performance of black-box diffusion policies, without changing the base parameters, making it lightweight and flexible.
We present results from both simulation and real-world data and achieve an average 49\% improvement in success rate over the base policy across 5 real tasks. 
\end{abstract}

\IEEEpeerreviewmaketitle

\section{Introduction}

Behavior Cloning (BC) has become the de-facto standard for training manipulation policies from human teleop data~\cite{brohan2022rt, brohan2023rt, chi2023diffusionpolicy, black2024pi_0, kim2024openvla}.
Despite their popularity however, BC-based policies can be brittle and their performance varies wildly even within known tasks~\cite{mandlekar2021matters, ross2011reduction, wen2020fighting}.
Such failures of BC policies often come from imprecise actions on crucial fine-grained interaction points~\cite{florence2022implicit, mandlekar2020human, shafiullah2022behavior, zhao2023learning}. 
While these failures can be mitigated by collecting additional data, this requires laborious and costly data collection and curation~\cite{belkhale2023data, mandlekar2020human}, and human-collected data is not guaranteed to cover the policy's actual failure modes~\cite{kelly2019hg, laskey2017dart, ross2011reduction}.

To alleviate these issues, we use a rich source of data that is overlooked by other methods: the policy's own evaluation.
This data contains both successful demonstrations that could be used to reinforce what the policy does \textit{right} and also failures that provide valuable information about what the policy does \textit{wrong}. 
Intuitively, examples of failures contain crucial information about the task and environment, such as which states require especially precise control, or which strategies are error-prone. 



\begin{figure}[t!]
    \centering
    \includegraphics[width=\columnwidth]{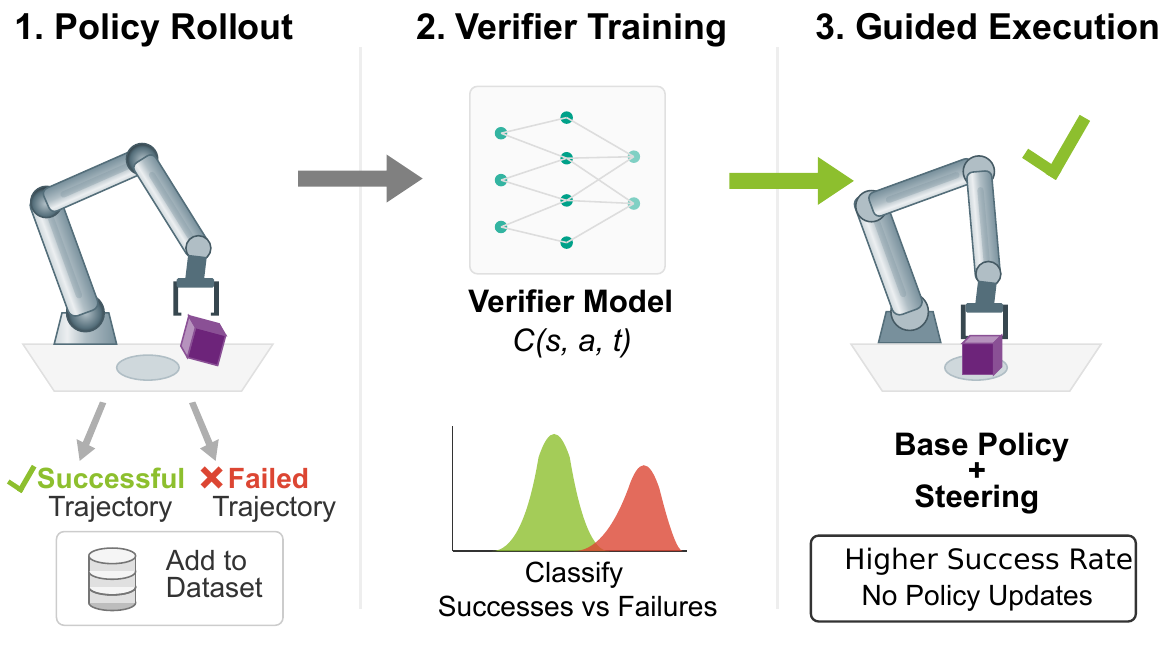}
    \caption{Our method relies on a policy's own evaluation data to improve its performance. Training small verifiers and subsequently, utilizing them via inference-time steering, allows for improved policy performance without costly data collections and resource extensive fine-tuning.}
    \label{fig:teaser_figure}
\end{figure}

To leverage this data, we propose Update-Free On-Policy Steering (\algoname) which uses data from a pre-trained BC policy's evaluations to \emph{steer}~\cite{ma2025inference, zhang2025inference} the policy without updating its parameters. 
The core of \algoname~is a trained verifier function, which predicts whether actions belong to a successful or failed rollout based on real evaluation data from the base policy.
This verifier function is then used to steer the action proposals of the base policies towards successful completion. 
To show the general applicability of our method, we present and implement two possible verifier training designs and two possible steering strategies. We consider a time-to-success estimator and a success classifier. For the policy improvement strategies, we use best-of-N and classifier guidance on the predicted mean action estimate, also known as forward universal guidance \cite{bansal2023universal}. 

In contrast to prior work on using steering to guide robot policies~\cite{du2025dynaguide, wang2025inference, wu2025foresight}, we focus on steering a policy using \textit{its own experience} i.e. trajectories collected by the policy itself, using both successful and failed trajectories. 
However, as opposed to RL-based methods such as DSRL~\cite{wagenmaker2025steering}, UF-OPS is able to achieve strong performance improvements without repeated online evaluation of the steered policy.
Instead, it is able to use a fixed dataset of on-policy data from the base-policy itself, which greatly simplifies the setup compared to online RL methods.
In addition, \algoname~does not require finetuning the base policy, making it applicable in compute-constrained or black-box scenarios, and mitigating the risk of catastrophic forgetting~\cite{wolczyk2024fine}.
Finally, \algoname~is fast and sample efficient in training and inference.

\begin{figure*}[h!]
\centering
\includegraphics[width=\textwidth]{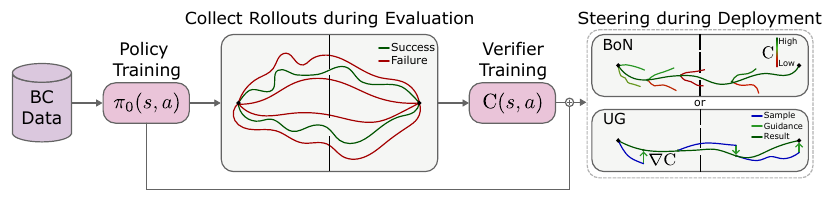}
\caption{Method overview: Given a base policy trained on a dataset of expert demonstrations, policy evaluation provides successful and failed rollouts. These are used for training a \textit{verifier function} that scores a transition $(s, a)$ in terms of its success likelihood. Finally, the verifier function is used in combination with a steering strategy, to improve the policy performance.}
\label{fig:method_overview}
\end{figure*}

We show that using a policy's evaluations leads to efficient steering, even with limited interactions.
In simulated manipulation tasks in the robomimic task suite \cite{mandlekar2021matters}, we find that \algoname~ is able to improve upon a base diffusion policy more reliably and with significantly smaller costs than prior methods.
In five real manipulation tasks on the Aloha system, we find that \algoname~increases success rates by 25 to 80 percentage points, a substantial improvement over the base diffusion policies.
In the real robot setup, it is able to achieve this improvement with as little as 60 evaluation trajectories per task.


\section{Background: Diffusion \& Flow Policies}


Behavioral Cloning~\cite{torabi2018behavioral} 
reproduces actions in a dataset containing expert demonstrations. 
Within this setup, a state-of-the-art approach is to represent the policy with a diffusion model~\cite{chi2023diffusionpolicy}. Diffusion models originate as image generation methods which deliver impressive high-fidelity output~\cite{dhariwal2021diffusion, ho2020denoising, song2020denoising}. 
They were adapted to behavior cloning as \emph{Diffusion Policies}~\cite{chi2023diffusionpolicy} and have become a standard design choice for manipulation systems.

Diffusion models enable sampling from a given distribution  $q(\boldsymbol{x}_0)$ matching the original training set (in our case, human demonstrations of robot actions). 
In order to do so, this distribution is approximated by a parameterized distribution $p_\theta(\boldsymbol{x}_0) = \int p_\theta(\boldsymbol{x}_{0:T})d\boldsymbol{x}_{1:T}$ with learned parameters $\theta$.
For this work, we utilize DDPM~\cite{ho2020denoising}, which models a diffusion process over individual samples in the data distribution as
\begin{equation}
\begin{split}
    q(\boldsymbol{x}_{1:T}|\boldsymbol{x}_0) &= \prod^T_{t=1}q(\boldsymbol{x}_t|\boldsymbol{x}_{t-1}), \\
    q(\boldsymbol{x}_{t}|\boldsymbol{x}_{t-1}) &\coloneqq \mathcal{N}(\boldsymbol{x}_t;\sqrt{1 - \beta_t}\boldsymbol{x}_{t-1}, \beta_t\boldsymbol{I})\ .
\end{split}
\end{equation}
Here $\beta_t$ represents the timestep-dependent level of noise added in the forward process. The model is then trained to predict the added noise. I.e. given a learned noise model $\boldsymbol{\epsilon}_\theta(\boldsymbol{x}_t, t)$, DDPM obtains a denoised sample via
\begin{equation}
\boldsymbol{x}_{t-1} = \frac{1}{\sqrt{\alpha_t}} \left( \boldsymbol{x}_t - \frac{1 - \alpha_t}{\sqrt{1 - \bar{\alpha}_t}} \boldsymbol{\epsilon}_\theta(\boldsymbol{x}_t, t) \right) + \sigma_t \boldsymbol{z}_t
\end{equation}
for each $t \in (T, .., 1)$, where $\alpha_t = 1 - \beta_t$, $\bar{\alpha}_t = \prod_{s=1}^t \alpha_s = \prod_{s=1}^t (1 - \beta_s)$, $\beta_t \in (0, 1)$ is the variance schedule, and $\boldsymbol{x}_T, \boldsymbol{z}_T, \ldots \boldsymbol{z}_1\sim \mathcal{N}(0, \boldsymbol{I})$ are random standard Gaussian noise.

To adapt diffusion models to the behavior cloning setting, the model is expanded by an additional conditional input, the state observation $\mathbf{s}$.
The diffused sample corresponds to the action that the robot is executing.
In addition, it has become common practice to generate a small sequence of $n$ actions in one go, a strategy commonly known as action chunking~\cite{zhao2023learning}.
For a sequence of actions $a^{[1:n]}$, the diffusion process becomes

\begin{align}
    \boldsymbol{a}^{[1:n]}_{t-1} = \frac{1}{\sqrt{\alpha_t}}\left(\boldsymbol{a}^{[1:n]}_t - \frac{1 - \alpha_t}{\sqrt{1 - \bar{\alpha}_t}} ~\boldsymbol{\epsilon}_\theta\left(\boldsymbol{a}^{[1:n]}_t, t\middle|s\right) \right) + \sigma_t \boldsymbol{z}
\end{align}

Instead of diffusion models, which rely on a stochastic differential equation to model the generative process, we can also use flow models \cite{albergo2023building,liu2023flow}.
A flow matching policy is defined by the ODE
\begin{equation}
    \frac{da}{dt} = v_t(a) \qquad a_t = (1 - t) \cdot a_0 + t \cdot a_1.
\end{equation}

Equivalent to a diffusion model, flow models can be easily adapted to represent action distributions for policy training \cite{hu2024adaflow,braun2024riemannian,black2024pi_0}.
Due to the close mathematical connections between diffusion and flow policies, the guidance strategies discussed in this paper transfer cleanly between both, and we evaluate models using both approaches in the experimental section.

\textbf{Single-step estimation of clean samples.}
To apply some steering techniques, it is necessary to predict a clean sample $\hat{a}_0$ from an intermediate noisy one.
This can be achieved at each timestep of the ODE via the Tweedie formula~\cite{efron2011tweedie,kim2021noise2score}
\begin{equation}
        \hat{\mathbf{a}}^t_0 \gets \frac{1}{\sqrt{\bar{\alpha}_k}} \left( \mathbf{a}^t_k - \sqrt{1-\bar{\alpha}_k} \mathbf{\epsilon} \right). \label{eq:tweedie}
\end{equation}
In flow models, this simplifies to
\begin{equation}
    \hat{a}_0^t = a_t - t \cdot v_t.
\end{equation}


\section{Related Work}

Three broad strategies have emerged for adapting and improving a base policy.
First, demonstration data can be collected to further refine the policy using imitation or reinforcement learning~(\cref{sec:fine_tuning}).
Second, manual or automatic data curation can be used to improve base policy training and thus performance~(\cref{sec:data_curation}).
Finally, it is possible to steer a policy without modifying its weights~(\cref{sec:steering}).

\subsection{Fine-tuning for Self-improvement}\label{sec:fine_tuning}

Policy improvement via obtaining new data and fine-tuning is a well-established field~\cite{ross2011reduction, bousmalis2023robocat, jin2025sime, jin2025soe}.
The seminal method for improving imitation learning based on policy rollouts is DAgger~\cite{ross2011reduction} where an expert adds corrections to online rollouts. A more recent notable approach is RoboCat~\cite{bousmalis2023robocat}, which uses a self-improvement loop where a specialized fine-tuned agent is deployed to generate successful trajectories for a new task, which are then added back to the main dataset to train a more capable generalist agent in the next iteration.  

An alternative method for fine-tuning is reinforcement learning. While it has traditionally primarily been used for training policies from scratch, some recent work has shifted their attention to finetuning~\cite{hirose2024selfi, nair2020awac, ghasemipour2025self, raad2024scaling}. 
Our work is closest to \citet{ghasemipour2025self} where a success classifier is trained to predict task success given a state-action pair. This classifier is subsequently used as a reward model for fine-tuning a VLA on robot trajectories. In contrast, UF-OPS achieves high performance without updating the policy. 

\begin{figure*}[h]
\centering
\includegraphics[width=\textwidth]{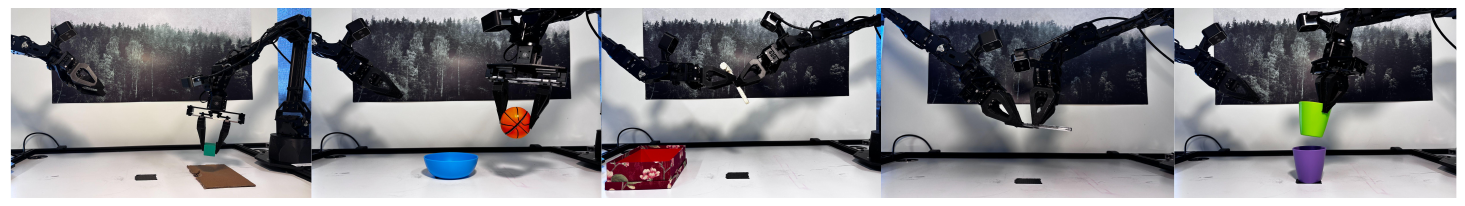}
\caption{Real tasks on the Aloha bimanual system~\cite{zhao2023learning} used for evaluation. From left to right, a) pick and place the block on the cardboard, b) pick and place the ball in the bowl, c) pick up the hammer with the right hand, hand it over to the left hand and drop it in the box, d) pick up the pen cap with the right hand and the cap with the left and insert it on the pen, and e) pick up the green cup to stack it on the purple one.}
\label{fig:real_tasks}
\end{figure*}

\subsection{Data Curation}
\label{sec:data_curation}

Another angle to address policy improvement is collecting or selecting higher quality data. For example, SEIL~\cite{jia2022seil} proposes augmenting expert trajectories with new simulated transitions using an equivariant model that explots symmetries in $SO(2)$. SART~\cite{oh2025self} leverages a single demonstration with precision-boundary annotations followed by robot self-augmentation within these boundaries. Demo-SCORE~\cite{chen2025curating} proposes training a classifier on success and failure rollouts and subsequently using it to filter the base dataset for robust behavior. Our work is similar in spirit to DEMO-Score, however, UF-OPS avoids retraining the policy and instead uses the classifier to choose higher value actions at test time.

\subsection{Steering and Guidance for Policy Improvement}
\label{sec:steering}

Recently, steering or guiding a pretrained base model towards policy improvement has gained some traction~\cite{ajay2022conditional, frans2025diffusion, liu2024bidirectional, nakamoto2024steering, wagenmaker2025steering, wang2025inference, zhang2025inference}. DSRL~\cite{wagenmaker2025steering} proposes to steer diffusion policies by training an auxiliary policy via RL.
This auxiliary policy predicts an action in the input noise space of the frozen diffusion model.
CFGRL~\cite{frans2025diffusion} trains a diffusion policy with Classifier-Free Guidance using an advantage function. DynaGuide~\cite{du2025dynaguide} trains a goal-conditioned dynamics model for classifier guidance. Our work is closest to V-GPS~\cite{nakamoto2024steering}, which proposes steering via a Q function trained on a large general-purpose dataset with offline RL, in order to improve base VLA performance. Unlike V-GPS, we focus on \textit{self-improvement} and use relatively small, targeted datasets easily obtained from the policy itself. On-policy data allows us to sidestep the brittle and hard-to-tune pessimism parameter that is inherent to offline RL methods such as offline Q learning.

\section{Method: Steering with evaluation rollouts}
When BC policies are evaluated, large amounts of \textit{on-policy data} are generated.
However, this low-cost data is often underutilized, especially in real robotics settings, despite containing information on the policies' specific \textit{failure} modes. 
Therefore, \algoname~uses this on-policy data containing both successes and failures by following a general four-step framework: 
\begin{tcolorbox}
\begin{enumerate}[noitemsep,leftmargin=1em]
    \item Train a BC policy $\pi_0(\mathbf{a}|\mathbf{s})$ on an expert dataset.
    \item Collect successful and failing trajectories.
    \item Train a verifier $C(\mathbf{s},\mathbf{a}, t)$ to assign an action score.
    \item Use the verifier to \emph{steer} the policy towards a sample with higher score.
\end{enumerate}
\end{tcolorbox}

This approach combines the strengths of verifier-guided methods like V-GPS (avoiding expensive updates to the policy) with the advantages of reinforcement learning approaches (self-improvement from on-policy data).
This framework can also be interpreted as one step of policy iteration~\cite{sutton1998rl}.
However, when using online, on-policy data with binary rewards, Q-learning simplifies to success prediction and common issues in RL such as overestimation do not arise.

We first present the setup and notation (\cref{sub:setup}). Next, we discuss example verifiers (\cref{sub:verifiers}) and finally, we will showcase example guidance methods (\cref{sub:steering_methods}).

\subsection{Setup}
\label{sub:setup}
We start with a base policy $\pi_0(\mathbf{a}|\mathbf{s})$ for a given task. The policy undergoes standard evaluation and a new dataset of $N$ rollouts is collected: $D' = \{\tau_1, \dots, \tau_N \}$ where  $\tau_n = (\{\mathbf{s_0}, \mathbf{a_0}, \dots,\mathbf{s_T},\mathbf{a_T}\}, r_n)$ is the $n$-th trajectory, $\mathbf{s_i}$ and $\mathbf{a_i}$ are the $i$-th state and action in the trajectory, respectively, and $r_n$ is the binary success signal of the $n$-th trajectory. To avoid complex dense labeling of trajectories, we limit ourselves to the sparse binary signal of final episode success, which is often already collected as part of performance calculation.

\subsection{Example Verifier Methods}
\label{sub:verifiers}


For this work, we focus on light-weight and small verifiers trained with a small amount of rollouts, as collecting many policy evaluations can be time-consuming in a real world setting.
We explore two related options for verifier design, namely success prediction and time-to-success prediction.
For both, we expand each sample in the trajectory data with binary success labels into state-action pairs combined with the corresponding timestep in the episode, $(\mathbf{s}_t,\mathbf{a_t}, t)$ in the form of a sinusoidal embedding. 

The sinusoidal embedding used is calculated by: 

\begin{equation}
\label{eq:sinus_emb}
\text{Emb}(t, j) = 
\begin{cases} 
\sin\left(\frac{t}{10000^{\frac{j}{d}}}\right) & \text{if } j \text{ is even,} \\
\cos\left(\frac{t}{10000^{\frac{j-1}{d}}}\right) & \text{if } j \text{ is odd.}
\end{cases}
\end{equation}

\textbf{Success Classification}. 
We can then directly train a binary classifier $C(\mathbf{s}_t,\mathbf{a_t}, t)$ that predicts whether a given such tuple belongs to a successful episode or not on this data.

\textbf{Time-To-Success Estimation}.
To obtain a time-to-success predictor, we adapt the standard exponential discounting scheme from RL.
If we use the final-state success label as a sparse reward, the Q function at each timestep $t$ is equal to the exponentially discounted time-to-go of the trajectory:
\begin{equation}
\label{eq:mc_q}
    Q(\mathbf{s}_t, \mathbf{a}_t, t) = \gamma^{T-t} \cdot r_{T}\ ,
\end{equation}
where $T$ is the timestep of success in the trajectory, $r_T$ is the reward on that timestep, and $\gamma$ is a discount factor. 
This forces the verifier to encode not only whether an action is likely to lead to success, but it will also allow it to disambiguate which action leads to success faster.
Intuitively, this provides a more fine-grained signal to the policy improvement method.
Since we are dealing solely with on-policy data, we can forgo methods such as bootstrapping, which greatly simplifies and stabilizes training and the interpretation of the method. 

\textbf{Loss functions.} For the success classifier, we use a standard Binary Cross Entropy loss $\mathcal{L}_\text{BCE}$. However, in practice, we found it helpful to further regularize the representation of the classifier by adding an auxiliary contrastive loss~\cite{hadsell2006dimensionality}.
For the contrastive loss calculation, each positive sample from a successful trajectory $\mathbf{s^+},\mathbf{a^+}$ is paired with a negative sample from a failed trajectory $\mathbf{s^-},\mathbf{a^-}$, and vice versa.
The auxiliary loss is the $L_2$ distance between the embeddings, 
\begin{equation}
\label{eq:aux_loss}
\begin{split}
    \mathcal{L}_\text{aux} =  \max\big(0, \epsilon - \|\mathbf{z}(\mathbf{s^+},\mathbf{a^+},t^+)
    - \mathbf{z}(\mathbf{s^-},\mathbf{a^-},t^-)\|\big)^2,
\end{split}
\end{equation}
where $\epsilon$ is a similarity threshold and $\mathbf{z(s, a, t)}$ is the penultimate layer of the classifier.


The full loss used to train the classifier is 
\begin{equation}
\label{eq:full_loss}
    \mathcal{L} = \mathcal{L}_\text{BCE} + \lambda_\text{aux} \cdot \mathcal{L}_\text{aux}
\end{equation}

where $\lambda_{aux}$ is a weight hyperparameter. 

For the time-to-success estimation, we fit the targets using least squares regression. 

\begin{equation}
\label{eq:mse_loss}
    \mathcal{L} = \mathcal{L}_\text{MSE} = ||Q(\mathbf{s}_t, \mathbf{a}_t, t) - G_t(\mathbf{s}_t, \mathbf{a}_t, t)||^2
\end{equation}

where $G_t(\mathbf{s}_t, \mathbf{a}_t, t)$ is the exponentially discounted time-to-go from \autoref{eq:mc_q}.

\subsection{Example Steering and Guidance Strategies}
\label{sub:steering_methods}

Given a good verifier function, there are multiple choices for guidance strategies. We explore two variants.

\textbf{Inference-time Action Selection}. The simplest way to use a verifier is to generate and rank multiple action candidates~\cite{ma2025inference, nakamoto2024steering, wang2025inference, zhang2025inference}. Following~\citet{nakamoto2024steering} we adopt a simple best-of-N strategy with a greedy argmax~(\autoref{eq:best_of_N_argmax}). 

\begin{equation}
\label{eq:best_of_N_argmax}
    a_\mathrm{selected} = \argmax_{\mathbf{a_t} \in \mathcal{A}_\mathrm{prop}}C(\mathbf{s_t},\mathbf{a_t}, t)
\end{equation}

where $\mathcal{A}_\mathrm{prop}$ is the set of action proposals. Pseudo-code for the action selection algorithm can be found in Algorithm \ref{alg:boN_steering}. 
For steering with Best-of-N, we require sufficient stochasticity to ensure diverse action proposals. 
In some cases, single-task Diffusion Policies can become almost deterministic.
This effectively collapses a Best-of-N strategy to a Best-of-1. 
However, we do not observe such determinism in our multitask policy experiments, where we use the base model as-is. 
For the single task experiments, we stop training early to obtain a stochastic, yet potentially sub-par policy. 
This experimental design decision is made as our goal is not to achieve state-of-the-art performance but rather to evaluate whether our method can improve upon a suboptimal policy.

\begin{algorithm}[t]
\caption{Verifier Best-of-N Steering}
\label{alg:boN_steering}
\begin{algorithmic}[1]
    \Require Policy $\pi_0(s, a)$, verifier function $C(s, a, t)$, number of action samples $N$, max steps $T$ for a rollout
    \State Initialize $t \gets 0$, $s_0$
    \For{$t=$ 0 to $T$}
        \State Sample \{$\mathbf{a}_{t_0}, \mathbf{a}_{t_1}, ..., \mathbf{a}_{t_{N-1}}$\} from $\pi_0$

        \State $\mathcal{A}_t \gets \{\mathbf{a}_{t,0}, \dots, \mathbf{a}_{t,N-1}\}$
        \State $\mathbf{a}_t \gets \text{argmax}_{\mathbf{a} \in \mathcal{A}_t} C(\mathbf{s}_t, \mathbf{a}, t)$
    \EndFor
\end{algorithmic}
\end{algorithm}

\textbf{Classifier Guidance}. 
Instead of selecting the best option among a number of candidates, we can perturb the process generating the action using the verifier as an energy-based model~\cite{du2019implicit}.
The most common framework in the context of diffusion models is classifier-guided sampling~\citep{dhariwal2021diffusion}.

However, standard classifier guidance (CG) requires training the verifier to distinguish actions at all noise levels.
In practice, training with noised actions led to a verifier which ignores the action entirely and predicts success solely from the state, which makes the verifier unusable for steering.
We are thus limited to verifier functions trained only on unmodified observation-action pairs as obtained by the rollouts.
Given that actions are still noisy during denoising, to avoid a distribution shift between training and inference of the verifier, we adopt the Forward Universal Guidance \cite{bansal2023universal} variant of CG and recover estimated clean actions via the Tweedie formula (\autoref{eq:tweedie}).

In CG, the classifier forward pass uses the mean prediction of the final clean sample $\mathbf{\hat a}_0$ which DDPM approximates at every step of the reverse process. The gradient of the classifier w.r.t. $\mathbf{\hat a}_0$, $\nabla_{\mathbf{\hat a_0}} C(\mathbf{s_t}, \mathbf{\hat a_0}, t)$ is then multiplied by a strength guidance $\lambda $ and added as perturbation to $\mathbf{\hat a}_0$ and the remainder of the denoising step uses the new perturbed mean $\mathbf{\hat a}_0$. We include classifier guidance as a DDPM modification in the context of our method in Algorithm~\ref{alg:guided_ddpm_explicit_x0_modified}.

\begin{algorithm}[t]
\caption{Verifier Classifier-Guidance with DDPM}
\label{alg:guided_ddpm_explicit_x0_modified}
\begin{algorithmic}[1]
    \Require Diffusion policy $\pi_0(s, a)$ trained as a noise predictor $\mathbf{\epsilon}_\theta$, verifier function $C(s, a, t)$, gradient scale $s$, guidance strength $\lambda$, max steps $T$ for a rollout, $K$ diffusion steps.
    \State Initialize $t \gets 0, s_0$ initial sim/real state.
    \State Initialize noise schedule parameters $\alpha_k, \bar{\alpha}_k, \sigma_k^2$ for $k=1 \dots K$.
    \For {$t$ from 0 to $T$}
    \State Sample $\mathbf{a}_K^0 \sim \mathcal{N}(\mathbf{0}, \mathbf{I})$
    \For{$k$ from $K$ down to 1}
        \State $\mathbf{z} \sim \mathcal{N}(\mathbf{0}, \mathbf{I})$ if $k > 1$, else $\mathbf{z} = \mathbf{0}$
        
        \State $\mathbf{\epsilon} \gets \mathbf{\epsilon}_\theta(\mathbf{s}_t, \mathbf{a}^t_k, k)$
        \State $\hat{\mathbf{a}}^t_0 \gets \frac{1}{\sqrt{\bar{\alpha}_k}} \left( \mathbf{a}^t_k - \sqrt{1-\bar{\alpha}_k} \mathbf{\epsilon} \right)$
        \State $\hat{\mathbf{a}}^t_0 \gets \hat{\mathbf{a}}^t_0 \textcolor{red}{ +  \lambda \nabla_{\mathbf{\hat a}_0} \log C(\mathbf{s}_t, \hat{\mathbf{a}}^t_0, t)}$
        
        \State $\tilde{\mathbf{\mu}}_k \gets \frac{\sqrt{\bar{\alpha}_{k-1}}\beta_k}{1-\bar{\alpha}_k} \mathbf{a}^t_k + \frac{\sqrt{\alpha_k}(1-\bar{\alpha}_{k-1})}{1-\bar{\alpha}_k} \hat{\mathbf{a}}^t_0$
        
        \State $\mathbf{a}^t_{k-1} \gets \tilde{\mathbf{\mu}}_k + \sigma_k \mathbf{z}$
    \EndFor
    \EndFor
    \State \Return $\mathbf{a}^t_0$
\end{algorithmic}
\end{algorithm}


\definecolor{tableblue}{RGB}{235, 245, 255} 

\section{Experiments}


We first highlight the core principles of policy steering in a pedagogical navigation task~(\cref{sub:pedagogy}).
We further validate our method on single task policies~(\cref{sub:robomimic_exps}), where we demonstrate the importance of on-policy data for the efficacy of the approach~(\cref{sub:on_vs_off_policy}), and a multi task policy~(\cref{sub:libero_exps}), both in simulation, as well as on real world single task policies~(\cref{sub:real_exps}).

\begin{figure*}[t]
\begin{subfigure}[t]{0.7\linewidth}
\includegraphics[width=\linewidth]{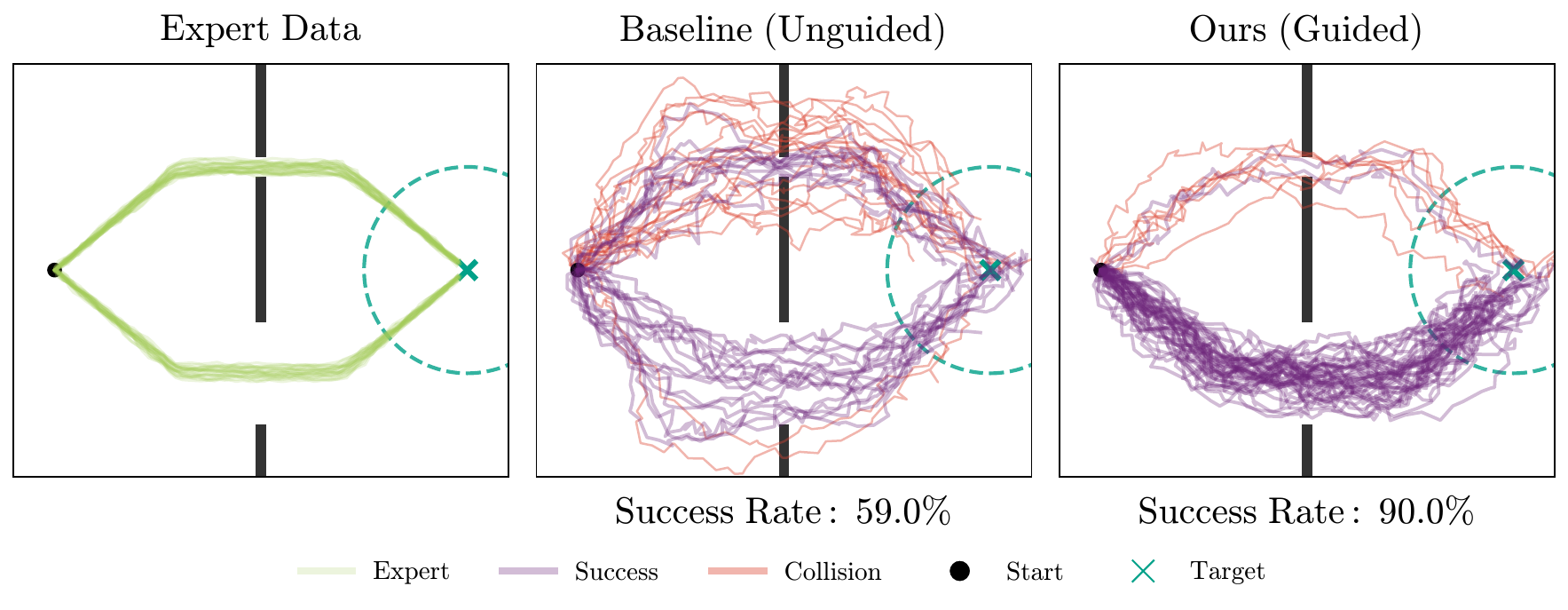}
    \subcaption{The expert demonstrations follow curves that go through both gaps. The unguided baseline frequently fails at the narrow gap. UF-OPS redirects traffic favoring the wide gap.}
    \label{fig:toy_example}
    \end{subfigure}
    ~
    \begin{subfigure}[t]{0.28\linewidth}
    \raisebox{16pt}{\includegraphics[width=\linewidth]{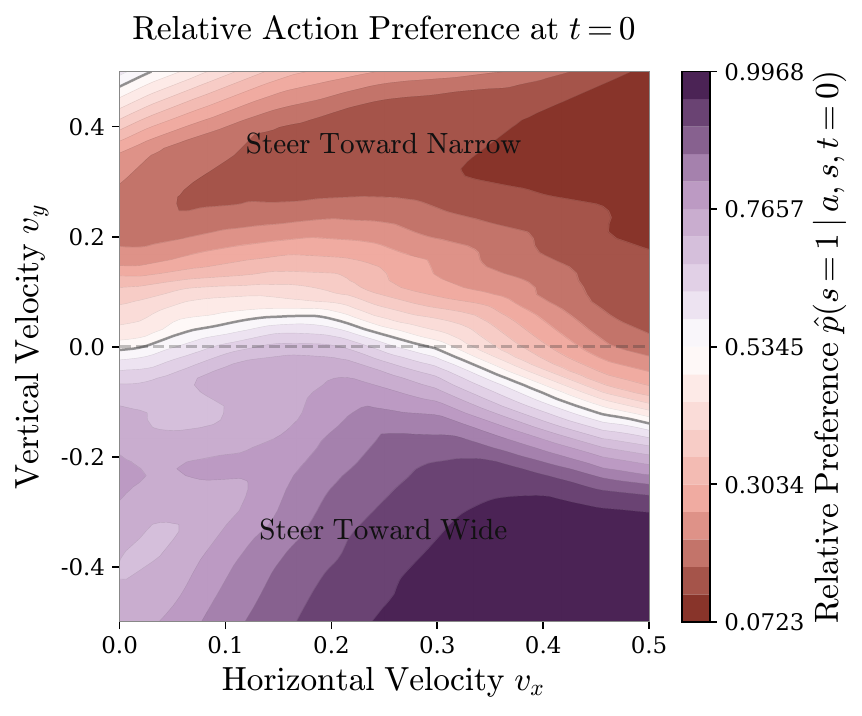}}
    \subcaption{The classifier identifies the wide gap (purple region) as the safer option.}
    \label{fig:toy_classifier_landscape}
    \end{subfigure}
    \caption{2D Pedagogical Example.}
\end{figure*}
\label{sec:method}

%

\subsection{Pedagogical Navigation Task}
\label{sub:pedagogy}
The pedagogical experiment is designed to visualize what kinds of failures an imitation policy might make when trained on expert data.
While an easy tasks such as this one can be perfectly solved by a powerful enough model, the intention of this example is not to benchmark performance improvements, but to highlight how steering can push a trained policy towards more robust modes of the policy space. 

\textbf{Setup}. The goal of this task is to move from the start point~(left) to the green circle on the right without colliding with a wall~(\autoref{fig:toy_example}). 
The agent can only control its velocity.
Any path 
must pass through one of the two gaps. Using the wide gap is easier as it requires less precision. 

The expert data is constructed from successful trajectories traversing either of the two gaps in equal proportions. 
In order to create a suboptimal policy and verify if our method can improve it, we emulate real-world effects, like actuator noise, slipping, or unmodeled dynamics, via adding some small environment noise in the agent’s position at test time.
Evaluating the resulting base diffusion policy on 1000 rollouts leads to a success rate of 59\%.
The majority of failures, shown in \autoref{fig:toy_example} are attempts to traverse the narrow gap.

\textbf{Verifier steering}. The evaluation trajectories are subsequently used to train a simple MLP classifier. During rollout, we use the Best-of-N approach to select the highest scoring of 30 candidate actions. The steered policy is re-evaluated on 1000 rollouts and yields a success rate of 90\%. Expert demos and samples of generated trajectories from the unguided and guided policies respectively, are presented in \autoref{fig:toy_example}.

\textbf{Visualization}. To further investigate steering, we provide a visualization of the classifier to showcase its preference at the start of a trajectory in \autoref{fig:toy_classifier_landscape}. In the beginning of the rollout, the classifier scores transitions that belong to the distribution mode corresponding to the wide gap more highly. As a result, the guided policy is steered towards the wide gap. Furthermore the generated trajectories from the guided policy form much tighter clusters in both gaps. This increases the chance of success even for the narrow gap.  

\subsection{Single Task Policy Simulation Experiments}
\label{sub:robomimic_exps}

\begin{table*}
\centering
\caption{Comparison of Performance (\%) on Robomimic SIM tasks for the base policy, four combinations of our method, Q function with Best-of-N (Q x BoN), classifier with Best-of-N (C x BoN), Q function with classifier guidance (Q x CG), and classifier with classifier guidance (C x CG), and the three baselines, V-GPS~\cite{nakamoto2024steering}, SAILOR~\cite{jain2025smooth}, and DSRL~\cite{wagenmaker2025steering}.}
\label{tab:sim_results}
\rowcolors{2}{tableblue}{} 
\begin{tabular}{lc|cc|c cccc} 
\toprule
\textbf{Task} & \textbf{Base} & \textbf{SAILOR}~\cite{jain2025smooth} & \textbf{DSRL}~\cite{wagenmaker2025steering} & \textbf{V-GPS}~\cite{nakamoto2024steering}&  \textbf{Q $\times$ BoN} & \textbf{C $\times$ BoN} & \textbf{Q $\times$ CG} & \textbf{C $\times$ CG} \\
\midrule
Transport (low dim) & 56.6{\tiny$\pm$3.07} & - & 24.8{\tiny$\pm$9.0} & - & 59.6{\tiny$\pm$ 3.04} & 62.7{\tiny$\pm$3.00} & \textbf{66.9{\tiny$\pm$2.92}} & 64.7{\tiny$\pm$3.0} \\
Square (low dim) & 78.2{\tiny$\pm$2.56} & - & 74{\tiny$\pm$6.1} & - & 85.1{\tiny$\pm$2.2} & \textbf{86.0{\tiny$\pm$2.2}} & 81.7{\tiny$\pm$2.4} & 85.5{\tiny$\pm$2.2} \\
\midrule
\midrule
Transport (image) & 58.1{\tiny$\pm$3.06} & 5.9{\tiny$\pm$1.46} & - & - &  65.7{\tiny$\pm$2.94} & \textbf{71.9{\tiny$\pm$2.79}} & 62.5{\tiny$\pm$3.0} & 60.7{\tiny$\pm$3.03}\\
Square (image) & 70.1{\tiny$\pm$2.84} & 45.1{\tiny$\pm$3.08} & -& 53.2{\tiny$\pm$3.09}  & 75.9{\tiny$\pm$2.7} & \textbf{83.5{\tiny$\pm$2.3}} & 76.4{\tiny$\pm$2.63} & 77.6{\tiny$\pm$2.6} \\
\bottomrule
\end{tabular}
\end{table*}

\textbf{Environment}. Our experimental setup comprises of four simulation tasks from the Robomimic suite~\cite{mandlekar2021matters}, namely low dimensional state Transport and Square, and image-based Transport and Square. Transport involves bimanual hand over of a hammer, and Square requires performing insertion of a square prop into a peg. 

\textbf{Diffusion BC Policy}. For our experiments, we use the original Diffusion Policy implementation~\cite{chi2023diffusionpolicy}. For each of the four simulation environments, a base diffusion policy is trained using the absolute action multi-human (MH) Robomimic datasets~\cite{mandlekar2021matters}. 
Each of the Robomimic MH datasets contains 300 demonstrations from 6 human operators of varying proficiency. 
We use the default backbone configuration for Robomimic environments provided by~\citet{chi2023diffusionpolicy}. Noise generation follows a DDPM~\cite{ho2020denoising} scheduler with 100 timesteps.

The state observation encompasses two Robomimic camera images, top down and agent view, along with proprioception over two timesteps. 
The output is an action chunk of 8. For low dimensional state, the observation is used as is as input to the verifier function while for image-based policies, we use the frozen embedding from the vision encoder of the base policy. 

To prevent deterministic predictions and to preserve the variance necessary for steering, we train low dimensional policies for 350 epochs and image-based policies for 150 epochs. 
This reduces the performance slightly below reported numbers in prior work.
Therefore this experiment is not supporting a claim on how to achieve state-of-the-art performance, it merely shows that UF-OPS is capable of improving upon suboptimal base policies.
We show that UF-OPS is capable of improving fully trained VLA policies in the next experiment.

For training the verifier, we collect 6048 trajectories for the low dimensional environments and 12012 trajectories for the image environments.
Low dimensional square trajectories contain 50 steps, transport trajectories contain 88 steps, and both image square and image transport trajectories contain 63 steps. 

\textbf{Verifier training}. The verifier is trained using the success label provided by the simulator.  
To prevent validation set contamination, entire trajectories are split 80-20 between training and validation. Verifiers are trained for 200 epochs and the checkpoint with the lowest validation loss is chosen for guidance. For classifier guidance, in practice we apply the gradient of the time-to-success Q estimation intact but still take the log of the classifier. For steering with Best-of-N argmax, $N = 10$ suffices. For classifier guidance, we present results for the best guidance strength per task. 

\begin{table}[t]
\centering
\caption{Contrastive loss ablation. Best-of-N policy steering using classifiers trained with and without the contrastive loss.}
\label{tab:contrastive_ablation}
\rowcolors{2}{tableblue}{} 
\begin{tabular}{lc|cc}
\toprule
\textbf{Task} & \textbf{Base} & $\lambda=0.1$ & $\lambda=0.0$) \\
\midrule
Transport (low dim) & 56.6{\tiny$\pm$3.07} & \textbf{62.7{\tiny$\pm$3.00}} & 56.0{\tiny$\pm$3.08} \\
Square (low dim) & 78.2{\tiny$\pm$2.56} & \textbf{86.0{\tiny$\pm$2.2}} & 85.5{\tiny$\pm$2.2} \\
\bottomrule
\vspace{-3em}
\end{tabular}
\end{table}

\textbf{Baselines and results}. We compare our results against three recent works, V-GPS~\cite{nakamoto2024steering}, DSRL~\cite{wagenmaker2025steering} and SAILOR~\cite{jain2025smooth}.
To provide a fair comparison, we limit all methods to the same amount of data. 
The Q-function of V-GPS is trained on visual single-arm data, so we only compare its performance on square image, as transport image is a bimanual task. We further adapt our evaluation to score the first action of each sampled chunk, choose the highest-scoring action and replan at every step. 
\autoref{tab:sim_results} shows that our method surpasses DSRL and SAILOR for the same number of on-policy interactions, and the pretrained V-GPS baseline. 

Interestingly, while DSRL starts with a similar baseline performance as our policy, training first decreases the performance as the RL components struggle to fit their targets with limited data, and only starts surpassing the baseline on transport after an order of magnitude more samples than used here.
In addition, DSRL requires a diffusion model with significantly fewer denoising steps to make efficient training possible.
This further decreases performance of the underlying model, whereas our method is able to work with an expensive and capable DP base.

In addition, the training of DSRL and SAILOR is significantly slower than our method.
For comparison, collecting 6000 trajectories for the square task takes circa two and a half hours on a RTX3090 GPU, with the runtime mostly dominated by the diffusion model prediction.
Fully training DSRL for the same number of steps takes an additional 4 hours, while training our verifier functions only takes about 20 minutes. 


\textbf{Ablation}. For classifier training, we isolate the importance of the contrastive loss. To study its effect, we focus on the two Robomimic low dimensional environments, Transport and Square, and repeat our experiments with the same classifier architecture but with the contrastive loss disabled and Best-of-N as the steering strategy. Results are presented in \autoref{tab:contrastive_ablation}. Interestingly, for Square the contrastive loss does not have a real impact, however for Transport its absence renders the classifier ineffective which suggests that for some tasks encouraging further separation of positive and negative inputs in the embedding space is required. 

\subsection{Design Decision and Base Policy Analysis}

\noindent\textbf{(1) On-policy vs off-policy rollouts}
\label{sub:on_vs_off_policy}

Beyond base efficacy, the question arises how important \textit{on-policy} data is for UF-OPS steering. To answer it, we conduct a set of experiments that leverage rollouts of one policy to steer a different base policy. 

We train a second set of base diffusion policies for low-dimensional Transport and Square. For these, we use the absolute action proficient human (PH) Robomimic datasets~\cite{mandlekar2021matters}. 
For fairness, we keep all the hyperparameters of the base policy and verifier training the same.

Subsequently, we attempt to guide the MH policy using the verifiers derived from rollouts from the PH policy and vice versa, using both steering strategies. As expected, the base PH policies perform better than their MH counterparts as they are trained on fewer but higher-quality demonstrations. Therefore, the verifiers trained on rollouts from PH versus MH are off-policy with respect to one another. 

\renewcommand{\arraystretch}{1}
\begin{table} 
\centering
\caption{Comparison of steering via verifiers trained on off-policy, yet task-related, data reveals that leveraging on-policy rollouts is crucial for the efficacy of the method.}
\label{tab:ph_mh_ablation}
\rowcolors{2}{tableblue}{} 
\resizebox{\columnwidth}{!}{ 
\begin{tabular}{llcccccc} 
\toprule
\textbf{Task} & \textbf{Base} & \textbf{Q x BoN} & \textbf{C x BoN} & \textbf{Q x CG} & \textbf{C x CG}\\
\midrule
PH $\rightarrow$ MH Transport (low dim) & 56.6{\tiny$\pm$3.07} & 54.7\tiny{$\pm$3.09} & 54.3{\tiny$\pm$3.09} & 42.3{\tiny$\pm$3.06}& 13.6{\tiny$\pm$2.12}\\
PH $\rightarrow$ MH Square (low dim) & 78.2{\tiny$\pm$2.56} & 78.6{\tiny$\pm$2.54} & 79.8{\tiny$\pm$2.49} & 82.1{\tiny$\pm$2.37}& 56.4{\tiny$\pm$3.07}\\
\midrule
MH $\rightarrow$ PH Transport (low dim) & 81.1{\tiny$\pm$2.43} & 79.5{\tiny$\pm$2.5} & 79.4{\tiny$\pm$2.51} & 75.1{\tiny$\pm$2.68}& 27.9{\tiny$\pm$2.78}\\
MH $\rightarrow$ PH Square (low dim) & 90.7{\tiny$\pm$1.8} & 90.7{\tiny$\pm$1.8} & 93.7{\tiny$\pm$1.51} & 92.8{\tiny$\pm$1.6} & 89.4{\tiny$\pm$1.91} \\
\bottomrule
\end{tabular}
}
\end{table}

\renewcommand{\arraystretch}{1}
\begin{table} 
\centering
\caption{Steering a highly performant policy with an on-policy verifier still improves upon baseline.}
\label{tab:high_perf}
\rowcolors{1}{tableblue}{} 
\resizebox{\columnwidth}{!}{ 
\begin{tabular}{llcccccc} 
\toprule
\textbf{Task} & \textbf{Base} & \textbf{Q x BoN} & \textbf{C x BoN} & \textbf{Q x CG} & \textbf{C x CG}\\
\midrule
PH $\rightarrow$ PH Square (low dim) & 90.7{\tiny$\pm$1.8} & 94.3{\tiny$\pm$1.8} & 94.4{\tiny$\pm$1.43} & 93.4{\tiny$\pm$1.54} & 92.5{\tiny$\pm$1.63} \\
\bottomrule
\end{tabular}
}
\end{table}

Experimental results are presented in \autoref{tab:ph_mh_ablation}. As no variant clearly improves base performance, we conclude that access to on-policy rollouts is crucial for UF-OPS.
This highlights that data collected from other policies is not suitable for improving performance via steering.

\noindent\textbf{(2) High performance base policies}
\label{sub:high_perf_policies}

Our previous experiments showed improvements over mediocre policies where bigger gains could be achieved. But is training a competent verifier still possible when fewer failures are available, e.g. when a policy is already 90\% successful?  

For this experiment we use the Square low dim policy trained on PH data with baseline performance of 90.7\%. We train the two types of verifiers using the same data we collected for the earlier ablations, and steer using both Best-of-10 and CG with strength 0.1. Results are reported in \autoref{tab:high_perf}. All four verifier-steering strategy combinations show gains over the baseline which indicates that \algoname \ remains effective even the low failure regime. Of course, it should be noted that more on-policy failures can always be collected by rolling out the policy more times, if desired.

\subsection{VLA Multitask Simulation Experiments}
\label{sub:libero_exps}

\definecolor{tableblue}{RGB}{235, 245, 255} 
\begin{table*}
\centering
\small
\caption{Performance (\%) on 11 unseen Libero tasks: Comparison of the base policy against our method with time-to-success estimation and both steering alternatives.}
\label{tab:libero_sim_results}
\newcolumntype{Y}{>{\centering\arraybackslash}X}

\setlength{\tabcolsep}{2pt} 
\begin{tabularx}{\textwidth}{@{} l YYYYYYYYYYY | Y @{}}
\toprule
\textbf{Method} & \textbf{task12} & \textbf{task26} & \textbf{task28} & \textbf{task31} & \textbf{task38} & \textbf{task43} & \textbf{task59} & \textbf{task60} & \textbf{task64} & \textbf{task77} & \textbf{task79} & \textbf{Mean} \\
\midrule
\midrule
\rowcolor{tableblue} Base            & 9.9{\tiny$\pm$4.14}  & 5.9{\tiny$\pm$3.26}  & 41.2{\tiny$\pm$7} & 13.8{\tiny$\pm$4.78} & 31.3{\tiny$\pm$4.78} & 9.9{\tiny$\pm$4.14}  & 18.8{\tiny$\pm$5.41} & 49.5{\tiny$\pm$6.93} & 35.9{\tiny$\pm$6.65} & 30.4{\tiny$\pm$6.37} & 33.6{\tiny$\pm$6.55} & 25.5 \\

V-GPS~\cite{nakamoto2024steering} & 31.5{\tiny$\pm$6.44} & 1.5{\tiny$\pm$1.68} & 18.0{\tiny$\pm$5.32} & 6.5{\tiny$\pm$3.41} & 10.5{\tiny$\pm$4.25} & 2.0{\tiny$\pm$1.94} & 10.0{\tiny$\pm$4.16} & 41.5{\tiny$\pm$6.83} & 28.0{\tiny$\pm$6.22} & 13.5{\tiny$\pm$4.74} & 26.0{\tiny$\pm$6.08} & 17.2\\

\rowcolor{tableblue} Q $\times$ BoN  & 25.0{\tiny$\pm$6} & 19.5{\tiny$\pm$5.49} & 72.0{\tiny$\pm$6.22} & 38.0{\tiny$\pm$6.73} & 67.5{\tiny$\pm$6.49} & 35.5{\tiny$\pm$6.63} & 47.5{\tiny$\pm$6.92} & 90.5{\tiny$\pm$4.06} & 78.5{\tiny$\pm$5.69} & 62.0{\tiny$\pm$6.73} & \textbf{83.0{\tiny$\pm$5.21}} & 56.3 \\
Q $\times$ CG   & \textbf{78.0{\tiny$\pm$5.74}}    & \textbf{20.0{\tiny$\pm$5.54}}    & \textbf{81.0{\tiny$\pm$5.44}}    & \textbf{58.0{\tiny$\pm$6.84}}    & \textbf{73.0{\tiny$\pm$6.15}}   & \textbf{84.5{\tiny$\pm$5.02}}    & \textbf{77.5{\tiny$\pm$5.79}}    & \textbf{97.0{\tiny$\pm$2.36}}   & \textbf{86.5{\tiny$\pm$4.74}}    & \textbf{98.5{\tiny$\pm$1.5}}    & 74.0{\tiny$\pm$6.08}  & \textbf{75.3} \\
\bottomrule
\end{tabularx}

\vspace{4pt}
\scriptsize
\textbf{Tasks:} \textbf{12}: put the black bowl at the back on the plate. \textbf{26}: put the wine bottle in the bottom drawer. \textbf{28}: close the top drawer of the cabinet. \textbf{31}: put the black bowl on top of the cabinet. \textbf{38}: put the right moka pot on the stove. \textbf{43}: put the white bowl on top of the cabinet. \textbf{59}: pick up the tomato sauce and put it in the tray. \textbf{60}: pick up the black bowl and put it in tray. \textbf{64}: stack the bowls and place them in the tray. \textbf{77}: pick up book (back compartment). \textbf{79}: pick up book (left compartment).
\end{table*}

\textbf{Environment}. While training specialized policies for desired single tasks can be very useful, one emerging trend over the past years has been towards using Vision-Language-Action (VLA) foundation models. These models are capable of achieving some success out-of-the-box on many novel tasks, but often fall short in terms of reliability and consistency.
We therefore test whether UF-OPS can be used to improve a pre-trained VLA policy towards better performance on zero-shot tasks.
To do so, we apply our method on the Libero benchmark \cite{liu2023libero} using the $\pi_{0.5}$ policy. We use the checkpoint finetuned on 40 Libero tasks taken from the Spatial, Object, Goal, and Libero-10 suites, provided by Physical Intelligence \cite{black2024pi_0}. It should be noted that $\pi_{0.5}$ is a flow model, not a diffusion policy \cite{lipman2022flow}.
The remaining 90 Libero tasks serve as an out-of-distribution test set that the checkpoint demonstrates varying levels of performance. 
As our method is aimed at policy improvement, we choose 11 tasks where the $\pi_{0.5}$ checkpoint results in poor or mediocre success rate, specifically from $5.9\%$ to $49.5\%$. To train the verifier, we collect 3000 trajectories for each of the target tasks. 

\textbf{Verifier training}. For the multi-task setup, we require multi-task verifiers. To achieve this, we concatenate a MUSE language embedding \cite{lample2017unsupervised}, following~\citet{nakamoto2024steering}, with vision, proprioception, action and progress timestep embeddings from the VLA.
This setup allows us to support VLAs with two tower architectures that keep language and vision separate. For $\pi_{0.5}$, this architecture is somewhat redundant as the policy tokens contain language information, however ablating this design choice shows that including it can slightly improve performance. We conjecture that since the backbone features are not explicitly trained to support verifier predictions, having a secondary input stream helps the verifier training. 
We train the verifier for 200 epochs and use the checkpoint with the best validation loss on a held-out subset of the training data. We then perform 200 rollouts per task with Best-of-N with $N=10$ and classifier guidance with guidance strength $\lambda=0.1$. 

Results are provided in \autoref{tab:libero_sim_results}. 
Both steering strategies show significant improvements with classifier guidance being clearly superior. Namely, best-of-N showcases a +31.0\% increase while classifier guidance achieves a +49.8\% increase. 
We find that V-GPS cannot achieve a competitive performance.
This is likely caused by V-GPS relying on a generic pre-trained verifier function that has most likely not encountered the Libero task suite during training.
This strengthens our hypothesis further that it is important to design and train a verifier on data specific to the steered policy.
While a pre-trained verifier might provide good starting points for on-policy verifier fine-tuning, we leave this direction open for future work.

We also implement a simple multi-task variant of DSRL and compare it on the Libero tasks.
We found that it is able to match and outperform UF-OPS with fewer environment interactions.
However, UF-OPS is designed to be evaluated with a static dataset, while DSRL obtains improved policies during online training.
Therefore it is not surprising that, given enough data, DSRL will be able to achieve greater performance.
The relative speed advantage of UF-OPS is also slightly less significant, as the runtime for either algorithm is dominated by the inference time of the underlying VLA model.
Extending UF-OPS to an online setting where the verifier is repeatedly updated with on-policy data from the steered policy is left as an exciting avenue for future work.

\begin{table} 
\centering
\caption{Performance (\%) of \algoname \ as a function of dataset size on Libero using $\pi_{0.5}$.}
\label{tab:dataset_size_perf}
\rowcolors{1}{tableblue}{} 
\begin{tabular}{l|cccc} 
\toprule
\textbf{Dataset Size} & 1000 & 2000 & 3000\\
\textbf{Performance (\%)} & 65.0\% & 69.8\% & 75.3\%  \\
\bottomrule
\end{tabular}
\end{table}

\textbf{Method improvement with dataset size.} To investigate how performance changes as a function of the offline dataset size used to train the verifier, we repeat our Libero $\pi_{0.5}$ experiments using 1000, 2000 and 3000 samples respectively. All three experiments use the same base policy and classifier guidance with coefficient 0.1 similarly to \autoref{tab:libero_sim_results}. Results can be seen in \autoref{tab:dataset_size_perf}. As expected, performance of \algoname \ continues to improve when increasing the offline dataset size.

\subsection{Real experiments}
\label{sub:real_exps}

\begin{figure*}[t]
\centering
\definecolor{uoftpurple}{HTML}{6D247A}
\definecolor{uoftteal}{HTML}{00A189}
\definecolor{uoftorange}{HTML}{DC4633}

\begin{subfigure}[b]{0.49\textwidth}
\centering
\setlength{\abovecaptionskip}{2pt}   
\setlength{\belowcaptionskip}{0pt}   
\begin{tikzpicture}
\begin{axis}[
    ybar, width=\linewidth, height=2.7cm, bar width=7pt,
    enlarge x limits=0.13, ymin=0, ymax=125,
    axis lines*=left, ylabel={Success rate (\%)},
    symbolic x coords={block-pp,ball-bowl,transport,pen-cap,stack-cups},
    xtick=data,
    xticklabel style={font=\scriptsize , yshift=2pt},
    yticklabel style={font=\tiny},
    xtick style={draw=none}, ymajorgrids=true, grid style={gray!30},
    label style={font=\footnotesize},
    nodes near coords, nodes near coords style={font=\tiny, anchor=south, color=black},
    legend style={font=\scriptsize, at={(1,1.04)}, anchor=south east,
                  legend columns=-1, draw=none},
]
\addplot+[fill=uoftorange, draw=black] coordinates {
    (block-pp,40) (ball-bowl,55) (transport,30) (pen-cap,10) (stack-cups,55)};
\addlegendentry{Base}
\addplot+[fill=uoftteal, draw=black] coordinates {
    (block-pp,90) (ball-bowl,95) (transport,100) (pen-cap,35) (stack-cups,80)};
\addlegendentry{C\,$\times$\,BoN}
\addplot+[fill=uoftpurple, draw=black] coordinates {
    (block-pp,95) (ball-bowl,90) (transport,80) (pen-cap,90) (stack-cups,80)};
\addlegendentry{Q\,$\times$\,BoN}
\end{axis}
\end{tikzpicture}
\caption{Success rates per task.}
\label{fig:real_abs}
\end{subfigure}
\hfill
\begin{subfigure}[b]{0.49\textwidth}
\centering
\setlength{\abovecaptionskip}{2pt}   
\setlength{\belowcaptionskip}{0pt}   
\begin{tikzpicture}
\begin{axis}[
    ybar, width=\linewidth, height=2.7cm, bar width=9pt,
    enlarge x limits=0.08, ymin=-15, ymax=105,
    ytick={0,25,50,75,100},
    axis lines*=left, ylabel={Improvement\\over base (pp)},
    ylabel style={align=center},
    symbolic x coords={block-pp,ball-bowl,transport,pen-cap,stack-cups},
    xtick=data,
    xticklabel style={font=\scriptsize, yshift=2pt},
    yticklabel style={font=\tiny},
    xtick style={draw=none}, ymajorgrids=true, grid style={gray!30},
    label style={font=\footnotesize},
    legend style={font=\scriptsize, at={(1,1.04)}, anchor=south east,
                  legend columns=-1, draw=none},
    extra y ticks={0}, extra y tick labels={},
    extra y tick style={grid=major, grid style={black!70, thick}},
    error bars/error bar style={black, thin},
    error bars/error mark options={rotate=90, mark size=2pt, line width=0.6pt},
]
\addplot+[fill=uoftteal, draw=black,
    error bars/.cd, y dir=both, y explicit]
coordinates {
    (block-pp,    50)  += (0,19.50) -= (0,29.32)
    (ball-bowl,   40)  += (0,21.19) -= (0,26.73)
    (transport,   70)  += (0,15.45) -= (0,27.19)
    (pen-cap,     25)  += (0,22.88) -= (0,26.25)
    (stack-cups,  25)  += (0,23.97) -= (0,28.89)
};
\addlegendentry{C\,$\times$\,BoN}
\addplot+[fill=uoftpurple, draw=black,
    error bars/.cd, y dir=both, y explicit]
coordinates {
    (block-pp,    55)  += (0,18.58) -= (0,28.32)
    (ball-bowl,   35)  += (0,22.01) -= (0,27.79)
    (transport,   50)  += (0,19.52) -= (0,30.76)
    (pen-cap,     80)  += (0,10.20) -= (0,28.43)
    (stack-cups,  25)  += (0,23.97) -= (0,28.89)
};
\addlegendentry{Q\,$\times$\,BoN}
\end{axis}
\end{tikzpicture}
\caption{Improvement over base with 95\% Newcombe CIs.}
\label{fig:real_diff}
\end{subfigure}

\caption{Performance on 5 real Aloha tasks. (a) Absolute success rate. (b) Improvement of each guided variant over the base policy, with 95\% Newcombe confidence intervals for the difference of binomial proportions~\cite[Method 10]{Newcombe1998}.}
\label{fig:real}
\vspace{-0.7em}
\end{figure*}

To evaluate our method beyond simulation, we construct five real tasks on the Aloha bimanual system~\cite{zhao2023learning}. For the real setup, we used the standard Aloha stationary system provided by Trossen Robotics. The Aloha real setup is comprised by two ViperX follower arms that perform the tasks, and two WidowX leader arms used for data collection via puppeteering. The setup also has 4 RealSense D405 cameras, one center overhead, one worms eye at the center of the table, and two wrist cameras mounted on the follower arms. Our installation follows the standard Trossen documentation and scripts. However, for data collection, we opted to amend the Trossen provided scripts to allow for variable episode length. 
Our setup and list of tasks are presented in \autoref{fig:real_tasks}. 

\textbf{Diffusion Policies}. For each task, we collect 100 demonstrations. We limit initial object position variance to ensure some success among the small-scale single-task diffusion policies. We train each base policy for 80000 steps with an action chunk of 8 and the current observation only. Similarly to simulated experiments, the base policy model is an image-based U-Net~\cite{ronneberger2015u} with a ResNet18~\cite{he2015deep} backbone. We use a DDPM~\cite{song2020denoising} noise scheduler and 50 diffusion training and inference timesteps. We reduced the number of steps compared to simulation experiments to ensure faster execution on the real robot. The base policy uses the 3 out of the 4 cameras of the Aloha system as observation --- top down, left wrist, right wrist --- as well as the proprioceptive joint state. 

\textbf{Verifier training}. During evaluation, we set 750 max steps for the rollouts for all tasks except pen cap insertion, where we increased the max steps to 1300 to account for its complexity and long-horizon execution. 60 rollouts are saved during evaluation for training the verifiers.
The success label is provided by the experimenter during the initial evaluation.  
For each task, we train a time-to-success estimator and a classifier. We train the Q function for 200 epochs and the classifier for 400 epochs, and select checkpoints based on validation loss. It should be noted that the limited initial object position variance does not undermine the verifier, as it is an action-quality evaluator conditioned on observations, and thus action variance is what matters most.

\textbf{Results}. We perform twenty steered evaluations for each task, both for the base policy and for each verifier function, to ensure identical evaluation conditions.
For our real experiments, we choose Best-of-N with $N = 10$ as our steering mechanism, for its ease and simplicity. Results are presented in \autoref{fig:real}. UF-OPS increases performance of the base policy on all instances, with gains spanning from 25\% to 80\%. 
The Q function and classifier are both on par for most tasks except pen cap insertion. For this task, while the classifier still boosts performance, Q function steering strongly outperforms it. We hypothesize that the difference lies in the fact that pen cap insertion trajectories are longer than the other 4 tasks. Thus, time-to-success estimation is likely a better scoring function for long-horizon tasks. 

\begin{table*}[h]
\centering
\footnotesize
\caption{Hyperparameters of the base (unguided) diffusion policy for all four
experimental settings. All settings use absolute-pose actions with a 6D rotation
representation and the corresponding Robomimic dataset. The low-dimensional
settings train a separate policy per demonstration source (\textsc{ph} and
\textsc{mh}); the image settings use \textsc{mh} only. In the image settings the
observation encoder is trained end-to-end with the denoising network.}
\label{tab:base-policy}
\begin{tabular*}{\textwidth}{@{\extracolsep{\fill}}lcccclcccc@{}}
\toprule
& \multicolumn{2}{c}{\textbf{Low-dim.}} & \multicolumn{2}{c}{\textbf{Image}} & & \multicolumn{2}{c}{\textbf{Low-dim.}} & \multicolumn{2}{c}{\textbf{Image}} \\
\cmidrule(lr){2-3} \cmidrule(lr){4-5} \cmidrule(lr){7-8} \cmidrule(lr){9-10}
\textbf{Hyperparameter} & Square & Transport & Square & Transport & \textbf{Hyperparameter} & Square & Transport & Square & Transport \\
\midrule
\multicolumn{5}{l}{\emph{Problem setup}} & \multicolumn{5}{l}{\emph{Diffusion process}} \\
Demonstration source & \textsc{ph}, \textsc{mh} & \textsc{ph}, \textsc{mh} & \textsc{mh} & \textsc{mh} & Scheduler & \multicolumn{4}{c}{DDPM} \\
RGB camera views & --- & --- & 2 & 4 & Train / inference timesteps & \multicolumn{4}{c}{100 / 100} \\
Image resolution & --- & --- & \multicolumn{2}{c}{$3\times84\times84$} & $\beta$ schedule & \multicolumn{4}{c}{\texttt{squaredcos\_cap\_v2}} \\
Low-dim.\ obs.\ dim.\ per step & 23 & 59 & 9 & 18 & $\beta_{\mathrm{start}}$, $\beta_{\mathrm{end}}$ & \multicolumn{4}{c}{$10^{-4}$, $2\times10^{-2}$} \\
Cond.\ feature dim.\ per step & 23 & 59 & 1033 & 2066 & Variance type & \multicolumn{4}{c}{\texttt{fixed\_small}} \\
Action dim.\ per step $d_a$ & 10 & 20 & 10 & 20 & Prediction type & \multicolumn{4}{c}{$\epsilon$} \\
Prediction horizon $T$ & \multicolumn{4}{c}{16} & Sample clipping & \multicolumn{4}{c}{yes, range $1.0$} \\
Observation steps $T_o$ & \multicolumn{4}{c}{2} & \multicolumn{5}{l}{\emph{Optimization}} \\
Executed action steps $T_a$ & \multicolumn{4}{c}{8} & Optimizer & \multicolumn{4}{c}{AdamW} \\
Latency steps & \multicolumn{4}{c}{0} & Learning rate & \multicolumn{4}{c}{$10^{-4}$} \\
Rotation representation & \multicolumn{4}{c}{6D} & Betas & \multicolumn{4}{c}{$(0.95, 0.999)$} \\
\multicolumn{5}{l}{\emph{Observation encoder} (image only)} & Adam $\epsilon$ & \multicolumn{4}{c}{$10^{-8}$} \\
RGB backbone & --- & --- & \multicolumn{2}{c}{ResNet-18, from scratch} & Weight decay & \multicolumn{4}{c}{$10^{-6}$} \\
Share backbone across views & --- & --- & \multicolumn{2}{c}{no} & LR schedule & \multicolumn{4}{c}{cosine, 500 warmup steps} \\
Normalization layers & --- & --- & \multicolumn{2}{c}{GroupNorm} & Batch size & \multicolumn{2}{c}{256} & \multicolumn{2}{c}{64} \\
Crop size (random / center) & --- & --- & \multicolumn{2}{c}{$76\times76$} & Epochs & \multicolumn{2}{c}{500} & \multicolumn{2}{c}{8000} \\
ImageNet normalization & --- & --- & \multicolumn{2}{c}{yes} & Gradient accumulation & \multicolumn{4}{c}{1} \\
\multicolumn{5}{l}{\emph{Denoising network}} & Encoder frozen & --- & --- & \multicolumn{2}{c}{no} \\
Backbone & \multicolumn{4}{c}{\texttt{ConditionalUnet1D}} & EMA inv.\ gamma / power / max & \multicolumn{4}{c}{$1.0$ / $0.75$ / $0.9999$} \\
Down dims. & \multicolumn{2}{c}{$[256, 512, 1024]$} & \multicolumn{2}{c}{$[512, 1024, 2048]$} & Validation ratio & \multicolumn{4}{c}{0.02} \\
Diffusion step embed.\ dim. & \multicolumn{2}{c}{256} & \multicolumn{2}{c}{128} &  & \multicolumn{4}{c}{} \\
Kernel size & \multicolumn{4}{c}{5} & & & & & \\
GroupNorm groups & \multicolumn{4}{c}{8} & & & & & \\
FiLM cond.\ (scale + bias) & \multicolumn{4}{c}{yes} & & & & & \\
Observation conditioning & \multicolumn{4}{c}{global FiLM} & & & & & \\
\bottomrule
\end{tabular*}
\end{table*}

\section{Reproducibility}

To enable complete reproducibility, we provide a full description of the networks for base policy and verifier training, along with all hyperparameters used for base policy, verifier training and inference (\autoref{tab:base-policy} and \autoref{tab:guidance}). 

\subsection{Low dimensional experiments} 

The architecture used for the Q function of the low dimensional experiments is comprised by two 2 linear layer followed by ReLU encoders, one for observation and one for action, along with a sinusoidal embedding for the timestep of the transition within the episode. Those embeddings are concatenated and passed through an MLP containing 2 blocks of linear layer, layernorm, ReLU followed by a dropout and a final linear layer. For the contrastive classifier, the architecture is identical however the output of the last ReLU and before the dropout is used as the contrastive embedding and the final linear layer as the classification head.

\begin{table*}[t]
\centering
\footnotesize
\caption{Guidance-model, inference-time guidance, and evaluation hyperparameters
for all four settings. \textsc{t2s} (\textsc{Time2Success}) regresses the
discounted return and \textsc{con} (\textsc{Contrastive}) classifies success;
rows marked \textsc{t2s} or \textsc{con} apply to that model only. Both take the
flattened observation history, standardized with dataset statistics, concatenated
with the flattened action chunk ($d_{\mathrm{act}} = T \cdot d_a$) in the
policy's normalized space --- the raw state for low-dim.\
($d_{\mathrm{obs}} = T_o \cdot d_o$), the frozen observation-encoder output for
image. A single $\lambda = 0.1$ transfers across both low-dimensional tasks and
both guidance models, whereas each image setting requires its own.}
\label{tab:guidance}
\begin{tabular*}{\textwidth}{@{\extracolsep{\fill}}lcccclcccc@{}}
\toprule
& \multicolumn{2}{c}{\textbf{Low-dim.}} & \multicolumn{2}{c}{\textbf{Image}} & & \multicolumn{2}{c}{\textbf{Low-dim.}} & \multicolumn{2}{c}{\textbf{Image}} \\
\cmidrule(lr){2-3} \cmidrule(lr){4-5} \cmidrule(lr){7-8} \cmidrule(lr){9-10}
\textbf{Hyperparameter} & Square & Transport & Square & Transport & \textbf{Hyperparameter} & Square & Transport & Square & Transport \\
\midrule
\multicolumn{5}{l}{\emph{Guidance model: inputs}} & \multicolumn{5}{l}{\emph{Guidance model: optimization}} \\
Observation representation & \multicolumn{2}{c}{raw state} & \multicolumn{2}{c}{encoder features} & Optimizer & \multicolumn{4}{c}{AdamW} \\
Obs.\ input dim.\ $d_{\mathrm{obs}}$ & 46 & 118 & 2066 & 4132 & Learning rate & \multicolumn{2}{c}{$10^{-3}$} & \multicolumn{2}{c}{$5\times10^{-5}$} \\
Action input dim.\ $d_{\mathrm{act}}$ & 160 & 320 & 160 & 320 & Weight decay & \multicolumn{2}{c}{$10^{-2}$} & \multicolumn{2}{c}{$10^{-4}$} \\
Progress-time input & \multicolumn{4}{c}{env.\ step index} & Betas & \multicolumn{4}{c}{$(0.9, 0.999)$} \\
\multicolumn{5}{l}{\emph{Guidance model: architecture}} & Adam $\epsilon$ & \multicolumn{4}{c}{$10^{-8}$} \\
Encoder width & 64 & 64 & 256 & 64 & LR schedule & \multicolumn{4}{c}{linear, $1.0 \to 0.01$} \\
Progress-time embed.\ dim. & 64 & 64 & 128 & 64 & Batch size & \multicolumn{4}{c}{1024} \\
Trunk hidden dim.\ 1 & 128 & 128 & 512 & 128 & Epochs & \multicolumn{4}{c}{200} \\
Trunk hidden dim.\ 2 & 64 & 64 & 256 & 64 & Train / val split & \multicolumn{4}{c}{0.8 / 0.2, by trajectory} \\
Progress-time embedding & \multicolumn{4}{c}{sinusoidal} & Model selection & \multicolumn{4}{c}{best val.\ loss} \\
Nonlinearity & \multicolumn{2}{c}{ReLU} & \multicolumn{2}{c}{GELU} &  & \multicolumn{4}{c}{} \\
Spectral normalization & \multicolumn{2}{c}{no} & \multicolumn{2}{c}{yes} & \multicolumn{5}{l}{\emph{Inference: best-of-$N$}} \\
LayerNorm in encoders & \multicolumn{2}{c}{no} & \multicolumn{2}{c}{yes} & Candidates $N$ & \multicolumn{4}{c}{10} \\
LayerNorm in trunk & \multicolumn{4}{c}{yes} & Score, \textsc{con} & \multicolumn{4}{c}{$\sigma(\mathrm{logit})$} \\
\textsc{t2s} obs.\ LayerNorm & \multicolumn{2}{c}{no} & \multicolumn{2}{c}{yes} & Score, \textsc{t2s} & \multicolumn{4}{c}{predicted value} \\
\textsc{t2s} obs.\ noise (train) & --- & --- & \multicolumn{2}{c}{0.02} & Selection rule & \multicolumn{4}{c}{$\arg\max$} \\
\textsc{t2s} dropout & \multicolumn{2}{c}{0.5 (head)} & \multicolumn{2}{c}{0.1 (trunk)} & Denoising steps / cand. & \multicolumn{4}{c}{100 (full DDPM)} \\
\textsc{con} dropout (head) & \multicolumn{4}{c}{0.5} & \multicolumn{5}{l}{\emph{Inference: classifier guidance}} \\
Output & \multicolumn{4}{c}{single scalar} & $\lambda$, \textsc{con} & 0.1 & 0.1 & 0.05 & 0.05 \\
\multicolumn{5}{l}{\emph{Guidance model: objective}} & $\lambda$, \textsc{t2s} & 0.1 & 0.1 & 0.5 & 0.8 \\
\textsc{t2s} loss & \multicolumn{4}{c}{MSE to disc.\ return} & Guidance target & \multicolumn{4}{c}{predicted $\hat{x}_0$} \\
\textsc{t2s} discount $\gamma$ & \multicolumn{4}{c}{0.99} & Gradient, \textsc{con} & \multicolumn{4}{c}{$\nabla \log \sigma(\mathrm{logit})$} \\
\textsc{t2s} reward & \multicolumn{4}{c}{sparse, $1$ at success} & Gradient, \textsc{t2s} & \multicolumn{4}{c}{$\nabla$ predicted value} \\
\textsc{t2s} grad.\ clipping & \multicolumn{4}{c}{$\| g \|_2 \le 1.0$} & Applied at & \multicolumn{4}{c}{every denoising step} \\
\textsc{con} loss & \multicolumn{4}{c}{BCE $+\ \lambda_c \mathcal{L}_{\mathrm{contr}}$} & \multicolumn{5}{l}{\emph{Evaluation protocol}} \\
\textsc{con} weight $\lambda_c$ & \multicolumn{4}{c}{0.1} & Evaluation episodes & \multicolumn{4}{c}{1000} \\
\textsc{con} margin & \multicolumn{4}{c}{1.0} & Max env.\ steps & 400 / 500$^{\dagger}$ & 700 & 500 & 700 \\
\textsc{con} label & \multicolumn{4}{c}{terminal success} & Policy weights & \multicolumn{4}{c}{EMA} \\
On-policy rollouts & \multicolumn{2}{c}{$6048$} & \multicolumn{2}{c}{$12012$} &  & \multicolumn{4}{c}{} \\
\bottomrule
\end{tabular*}

\vspace{2pt}
{\footnotesize $^{\dagger}$400 for \textsc{ph}, 500 for \textsc{mh}.}
\end{table*}



\subsection{Image experiments}

The architecture used for the Q function of the image-based experiments is comprised by 2 encoders, one for observation and one for action. Each encoder has a spectral norm applied on a linear layer followed by a layernorm, GELU, another sprectral norm over a linear layer and a layernorm. The observation is first passed by a layernorm and noise augmentation with std 0.02 at training time only, is added. Then the observation and action are passed by the respective encoders. The results are concatenated along with the sinusoidal embedding of the episode timestep of the transition and this input is passed through an MLP containing 2 blocks of a spectral norm over a linear layer, layernorm, ReLU with a dropout in between, followed by a final linear layer. The contrastive classifier has the same architecture with the only difference being that a dropout is used and the dropout along with the final linear layer play the role of the classifier head, where the output of the layers before the dropout serve as the embedding. 




\subsection{VLA experiments}

The verifier is a two-stream MLP that scores an (observation, action-chunk) pair
conditioned on language, proprioceptive state, and a scalar progress index.
Observations are the pre-computed, frozen $\pi_{0.5}$ PaliGemma prefix features
(the mean-pooled Gemma-2B prefix hidden state, 2048d), so no visual backbone
is trained. These features, the 512d MUSE language embedding, the 8d
proprioceptive state (end-effector position, quaternion, gripper) and a 64d
sinusoidal embedding of the episode-local query index $t$ are encoded by
Linear+ReLU layers to $256$, $64$, $64$ and $64$ dimensions respectively and
fused to a 256d context. The flattened 320d action chunk ($10 \times 32$)
is encoded to 256d in parallel, and the concatenation is mapped through
$512 \to 256 \to 128 \to 1$ with a sigmoid output, giving $Q \in (0,1)$.

We train with AdamW at a learning rate of $3 \times 10^{-4}$ and weight decay
$10^{-3}$, decayed linearly to $1\%$ of its initial value, with a batch size of
$256$ and gradients clipped to a global norm of $1.0$. Training runs for at most
$200$ epochs and stops early once the validation loss has not improved for $30$
of them. Returns are discounted at $\gamma = 0.99$ under a sparse reward that is
zero everywhere except the terminal step of a successful episode, where it is
$+1$. We draw $3000$ trajectories per task and hold out a fifth of
the episodes for validation, splitting at the episode level so that no episode
contributes to both halves.

At evaluation we roll out $200$ episodes per task on LIBERO-90, capping each
episode at $400$ environment steps and re-querying the policy every $5$ steps.
Every query integrates the flow ODE with $10$ Euler steps and clips the
resulting actions to $[-1, 1]$, while the verifier scores
$64 \times 64 \times 6$ stacked agentview and wrist observations rendered at
$256 \times 256$. Best-of-$N$ steering draws $K = 10$ candidate chunks per query
and executes the highest-scoring one, selecting greedily at temperature
$\beta = 0$; classifier guidance instead draws a single chunk and applies a
guidance strength of $c = 0.1$ at every ODE step after the first.


\section{Limitations}

Although this work presents a general framework for improving the performance of a base policy using existing evaluation data that would otherwise typically be unused, there are three main limitations. First, verifiers are trained on the target tasks and we therefore do not evaluate verifier generalization beyond the specific training tasks. Second, applying this work to real robots still maintains a small overhead of manual labeling of successful and failed rollouts. For the moderate amount of trajectories used in our experiments, this requires a human to supervise the experiment and immediately label the episode success. Third, steering methods can be sensitive to hyperparameter choice. Especially classifier guidance has been shown to be very sensitive to guidance strength. Tuning this hyperparameter per task can vary performance substantially, although we found that outside of extreme ranges, performance improves. Finally, tuning the guidance strength on a real robot potentially poses some safety risks as high values can lead to extreme actions. It is therefore adviseable to tune starting from low values. 

\section{Conclusion} 
\label{sec:conclusion}

UF-OPS is a novel framework for policy improvement at test-time without fine-tuning or expensive data collection requirements. 
Our approach relies on the policy's own experience, i.e., its own successful and failed rollouts, to train scoring function or verifier to act as a guidance model at sampling time.
The verifier functions are optimized to select better action samples or refine predicted samples using its gradient. Ultimately, we find that obtaining a good verifier leads to a consistent, low-cost performance improvement. 

\section*{Acknowledgments}

The authors would like to thank Dhruv Shah, Dushyant Rao, Florian Shkurti, Jonathan Kelly, and Jonathan Tompson for fruitful discussions and helpful feedback. Finally, the authors would like to extend a special thanks to Vikas Sindhwani and Carolina Parada for their continued support without which this work would not have been possible.


\footnotesize{
\bibliographystyle{IEEEtranN}
\bibliography{references}

@IEEEtranBSTCTL{IEEEexample:BSTcontrol,
  CTLuse_forced_etal       = "yes",
  CTLmax_names_forced_etal = "2",
  CTLnames_show_etal       = "1",
  CTLdash_repeated_names = "no"
}

@inproceedings{ajay2022conditional,
  title={Is conditional generative modeling all you need for decision-making?},
  author={Ajay, Anurag and Du, Yilun and Gupta, Abhi and Tenenbaum, Joshua and Jaakkola, Tommi and Agrawal, Pulkit},
  booktitle={Neural Information Processing Systems (NeurIPS)},
  year={2022}
}

@inproceedings{bansal2023universal,
  title={Universal guidance for diffusion models},
  author={Bansal, Arpit and Chu, Hong-Min and Schwarzschild, Avi and Sengupta, Soumyadip and Goldblum, Micah and Geiping, Jonas and Goldstein, Tom},
  booktitle={the IEEE/CVF conference on computer vision and pattern recognition},
  year={2023}
}

@article{belkhale2023data,
  title={Data quality in imitation learning},
  author={Belkhale, Suneel and Cui, Yuchen and Sadigh, Dorsa},
  journal={Neural Information Processing Systems (NeurIPS)},
  year={2023}
}

@article{black2024pi_0,
  title={$\pi_0 $: A Vision-Language-Action Flow Model for General Robot Control},
  author={Black, Kevin and Brown, Noah and Driess, Danny and Esmail, Adnan and Equi, Michael and Finn, Chelsea and Fusai, Niccolo and Groom, Lachy and Hausman, Karol and Ichter, Brian and others},
  journal={arXiv preprint arXiv:2410.24164},
  year={2024}
}

@inproceedings{brohan2022rt,
  title={Rt-1: Robotics transformer for real-world control at scale},
  author={Brohan, Anthony and Brown, Noah and Carbajal, Justice and Chebotar, Yevgen and Dabis, Joseph and Finn, Chelsea and Gopalakrishnan, Keerthana and Hausman, Karol and Herzog, Alex and Hsu, Jasmine and others},
  booktitle={Robotics: Science and Systems (RSS)},
  year={2023}
}

@inproceedings{brohan2023rt,
  title={Rt-2: Vision-language-action models transfer web knowledge to robotic control},
  author={Brohan, Anthony and Brown, Noah and Carbajal, Justice and Chebotar, Yevgen and Chen, Xi and Choromanski, Krzysztof and Ding, Tianli and Driess, D and Dubey, A and Finn, C and others},
  booktitle={Conference on robot learning (CoRL)},
  year={2023}
}

@inproceedings{bousmalis2023robocat,
  title={Robocat: A self-improving generalist agent for robotic manipulation},
  author={Bousmalis, Konstantinos and Vezzani, Giulia and Rao, Dushyant and Devin, Coline and Lee, Alex X and Bauz{\'a}, Maria and Davchev, Todor and Zhou, Yuxiang and Gupta, Agrim and Raju, Akhil and others},
  booktitle={International Conference on Learning Representations (ICLR)},
  year={2025}
}

@inproceedings{chen2025curating,
  title={Curating demonstrations using online experience},
  author={Chen, Annie S and Lessing, Alec M and Liu, Yuejiang and Finn, Chelsea},
  booktitle={Robotics: Science and Systems (RSS)},
  year={2026}
}

@inproceedings{chi2023diffusionpolicy,
	title={Diffusion Policy: Visuomotor Policy Learning via Action Diffusion},
	author={Chi, Cheng and Feng, Siyuan and Du, Yilun and Xu, Zhenjia and Cousineau, Eric and Burchfiel, Benjamin and Song, Shuran},
	booktitle={Robotics: Science and Systems (RSS)},
	year={2023}
}

@article{dhariwal2021diffusion,
  title={Diffusion models beat gans on image synthesis},
  author={Dhariwal, Prafulla and Nichol, Alexander},
  journal={Neural Information Processing Systems (NeurIPS)},
  year={2021}
}

@inproceedings{du2019implicit,
  title={Implicit generation and generalization in energy-based models},
  author={Du, Yilun and Mordatch, Igor},
  booktitle={Neural Information Processing Systems (NeurIPS)},
  year={2019}
}

@inproceedings{du2025dynaguide,
  title={DynaGuide: Steering Diffusion Polices with Active Dynamic Guidance},
  author={Du, Maximilian and Song, Shuran},
  booktitle={Neural Information Processing Systems (NeurIPS)},
  year={2025}
}

@inproceedings{florence2022implicit,
  title={Implicit behavioral cloning},
  author={Florence, Pete and Lynch, Corey and Zeng, Andy and Ramirez, Oscar A and Wahid, Ayzaan and Downs, Laura and Wong, Adrian and Lee, Johnny and Mordatch, Igor and Tompson, Jonathan},
  booktitle={Conference on Robot Learning (CoRL)},
  year={2022}
}

@article{frans2025diffusion,
  title={Diffusion Guidance Is a Controllable Policy Improvement Operator},
  author={Frans, Kevin and Park, Seohong and Abbeel, Pieter and Levine, Sergey},
  journal={arXiv preprint arXiv:2505.23458},
  year={2025}
}

@inproceedings{ghasemipour2025self,
  title={Self-Improving Embodied Foundation Models},
  author={Ghasemipour, Seyed Kamyar Seyed and Wahid, Ayzaan and Tompson, Jonathan and Sanketi, Pannag and Mordatch, Igor},
  booktitle={Neural Information Processing Systems (NeurIPS)},
  year={2025}
}

@INPROCEEDINGS{hadsell2006dimensionality,
  author={Hadsell, R. and Chopra, S. and LeCun, Y.},
  booktitle={IEEE Computer Society Conference on Computer Vision and Pattern Recognition (CVPR)}, 
  title={Dimensionality Reduction by Learning an Invariant Mapping}, 
  year={2006}
}

@INPROCEEDINGS{he2015deep,
  title={Deep residual learning for image recognition},
  author={He, Kaiming and Zhang, Xiangyu and Ren, Shaoqing and Sun, Jian},
  booktitle={IEEE Computer Society Conference on Computer Vision and Pattern Recognition (CVPR)}, 
  year={2016}
}

@inproceedings{hirose2024selfi,
  title={Selfi: Autonomous self-improvement with reinforcement learning for social navigation},
  author={Hirose, Noriaki and Shah, Dhruv and Stachowicz, Kyle and Sridhar, Ajay and Levine, Sergey},
  booktitle={Conference on Robot Learning (CoRL)},
  year={2024}
}

@inproceedings{ho2020denoising,
  title={Denoising diffusion probabilistic models},
  author={Ho, Jonathan and Jain, Ajay and Abbeel, Pieter},
  booktitle={Neural Information Processing Systems (NeurIPS)},
  year={2020}
}

@inproceedings{jain2025smooth,
  title={A Smooth Sea Never Made a Skilled $\text{SAILOR}$: Robust Imitation via Learning to Search},
  author={Jain, Arnav Kumar and Mohta, Vibhakar and Kim, Subin and Bhardwaj, Atiksh and Ren, Juntao and Feng, Yunhai and Choudhury, Sanjiban and Swamy, Gokul},
  booktitle={Neural Information Processing Systems (NeurIPS)},
  year={2025}
}

@inproceedings{jia2022seil,
  title={Seil: Simulation-augmented equivariant imitation learning},
  author={Jia, Mingxi and Wang, Dian and Su, Guanang and Klee, David and Zhu, Xupeng and Walters, Robin and Platt, Robert},
  booktitle={IEEE International Conference on Robotics and Automation (ICRA)},
  year={2023}
}

@inproceedings{jin2025sime,
  title={SIME: Enhancing Policy Self-Improvement with Modal-level Exploration},
  author={Jin, Yang and Lv, Jun and Yu, Wenye and Fang, Hongjie and Li, Yong-Lu and Lu, Cewu},
  booktitle={IEEE/RSJ International Conference on Intelligent Robots and Systems (IROS)},
  year={2025}
}

@inproceedings{jin2025soe,
  title={SOE: Sample-Efficient Robot Policy Self-Improvement via On-Manifold Exploration},
  author={Jin, Yang and Lv, Jun and Xue, Han and Chen, Wendi and Wen, Chuan and Lu, Cewu},
  booktitle={IEEE International Conference on Robotics and Automation (ICRA)},
  year={2026}
}

@inproceedings{kelly2019hg,
  title={Hg-dagger: Interactive imitation learning with human experts},
  author={Kelly, Michael and Sidrane, Chelsea and Driggs-Campbell, Katherine and Kochenderfer, Mykel J},
  booktitle={IEEE/RSJ International Conference on Robotics and Automation (ICRA)},
  year={2019},
}

@inproceedings{kim2024openvla,
  title={Openvla: An open-source vision-language-action model},
  author={Kim, Moo Jin and Pertsch, Karl and Karamcheti, Siddharth and Xiao, Ted and Balakrishna, Ashwin and Nair, Suraj and Rafailov, Rafael and Foster, Ethan and Lam, Grace and Sanketi, Pannag and others},
  booktitle={Conference on Robot Learning (CoRL)},
  year={2025}
}

@article{lample2017unsupervised,
  title={Unsupervised Machine Translation Using Monolingual Corpora Only},
  author={Lample, Guillaume and Conneau, Alexis and Denoyer, Ludovic and Ranzato, Marc'Aurelio},
  journal={arXiv preprint arXiv:1711.00043},
  year={2017}
}

@inproceedings{laskey2017dart,
  title={Dart: Noise injection for robust imitation learning},
  author={Laskey, Michael and Lee, Jonathan and Fox, Roy and Dragan, Anca and Goldberg, Ken},
  booktitle={Conference on Robot Learning (CoRL)},
  year={2017},
}

@inproceedings{lipman2022flow,
  title={Flow matching for generative modeling},
  author={Lipman, Yaron and Chen, Ricky TQ and Ben-Hamu, Heli and Nickel, Maximilian and Le, Matt},
  booktitle={International Conference on Learning Representations (ICLR))},
  year={2023}
}

@article{liu2023libero,
  title={Libero: Benchmarking knowledge transfer for lifelong robot learning},
  author={Liu, Bo and Zhu, Yifeng and Gao, Chongkai and Feng, Yihao and Liu, Qiang and Zhu, Yuke and Stone, Peter},
  journal={Advances in Neural Information Processing Systems (NeurIPS)},
  year={2023}
}

@inproceedings{liu2024bidirectional,
  title={Bidirectional Decoding: Improving Action Chunking via Guided Test-Time Sampling},
  author={Liu, Yuejiang and Hamid, Jubayer Ibn and Xie, Annie and Lee, Yoonho and Du, Maximilian and Finn, Chelsea},
  booktitle={International Conference on Learning Representations (ICLR)},
  year={2025}
}

@article{ma2025inference,
  title={Inference-time scaling for diffusion models beyond scaling denoising steps},
  author={Ma, Nanye and Tong, Shangyuan and Jia, Haolin and Hu, Hexiang and Su, Yu-Chuan and Zhang, Mingda and Yang, Xuan and Li, Yandong and Jaakkola, Tommi and Jia, Xuhui and others},
  journal={arXiv preprint arXiv:2501.09732},
  year={2025}
}

@article{mandlekar2020human,
  title={Human-in-the-loop imitation learning using remote teleoperation},
  author={Mandlekar, Ajay and Xu, Danfei and Mart{\'\i}n-Mart{\'\i}n, Roberto and Zhu, Yuke and Fei-Fei, Li and Savarese, Silvio},
  journal={arXiv preprint arXiv:2012.06733},
  year={2020}
}

@inproceedings{mandlekar2021matters,
  title={What matters in learning from offline human demonstrations for robot manipulation},
  author={Mandlekar, Ajay and Xu, Danfei and Wong, Josiah and Nasiriany, Soroush and Wang, Chen and Kulkarni, Rohun and Fei-Fei, Li and Savarese, Silvio and Zhu, Yuke and Mart{\'\i}n-Mart{\'\i}n, Roberto},
  booktitle={Neural Information Processing Systems (NeurIPS)},
  year={2021}
}

@inproceedings{nair2020awac,
  title={Awac: Accelerating online reinforcement learning with offline datasets},
  author={Nair, Ashvin and Gupta, Abhishek and Dalal, Murtaza and Levine, Sergey},
  booktitle={Deep Reinforcement Learning Workshop, Neural Information Processing Systems (NeurIPS)},
  year={2020}
}

@inproceedings{nakamoto2024steering,
  title={Steering your generalists: Improving robotic foundation models via value guidance},
  author={Nakamoto, Mitsuhiko and Mees, Oier and Kumar, Aviral and Levine, Sergey},
  booktitle={Conference on Robot Learning (CoRL)},
  year={2024}
}

@article{Newcombe1998,
  title = {Interval Estimation for the Difference between Independent Proportions: Comparison of Eleven Methods},
  shorttitle = {Interval Estimation for the Difference between Independent Proportions},
  author = {Newcombe, Robert G.},
  year = 1998,
  journal = {Statistics in Medicine},
  volume = {17},
  number = {8}
}

@article{oh2025self,
  title={Self-Augmented Robot Trajectory: Efficient Imitation Learning via Safe Self-augmentation with Demonstrator-annotated Precision},
  author={Oh, Hanbit and Murooka, Masaki and Motoda, Tomohiro and Nakajo, Ryoichi and Domae, Yukiyasu},
  journal={arXiv preprint arXiv:2509.09893},
  year={2025}
}

@article{raad2024scaling,
  title={Scaling instructable agents across many simulated worlds},
  author={Raad, Maria Abi and Ahuja, Arun and Barros, Catarina and Besse, Frederic and Bolt, Andrew and Bolton, Adrian and Brownfield, Bethanie and Buttimore, Gavin and Cant, Max and Chakera, Sarah and others},
  journal={arXiv preprint arXiv:2404.10179},
  year={2024}
}

@inproceedings{ronneberger2015u,
  title={U-net: Convolutional networks for biomedical image segmentation},
  author={Ronneberger, Olaf and Fischer, Philipp and Brox, Thomas},
  booktitle={Medical Image Computing and Computer-Assisted Intervention (MICCAI)},
  year={2015}
}

@inproceedings{ross2011reduction,
  title={A reduction of imitation learning and structured prediction to no-regret online learning},
  author={Ross, St{\'e}phane and Gordon, Geoffrey and Bagnell, Drew},
  booktitle={International Conference on Artificial Intelligence and Statistics (AISTATS)},
  year={2011},
}

@article{shafiullah2022behavior,
  title={Behavior transformers: Cloning $ k $ modes with one stone},
  author={Shafiullah, Nur Muhammad and Cui, Zichen and Altanzaya, Ariuntuya Arty and Pinto, Lerrel},
  journal={Neural Information Processing Systems (NeurIPS)},
  year={2022}
}

@inproceedings{song2020denoising,
  title={Denoising diffusion implicit models},
  author={Song, Jiaming and Meng, Chenlin and Ermon, Stefano},
  booktitle={International Conference on Learning Representations (ICLR)},
  year={2021}
}

@book{sutton1998rl,
  author    = {Sutton, Richard S. and Barto, Andrew G.},
  title     = {Reinforcement Learning: An Introduction},
  publisher = {MIT Press},
  year      = {1998}
}

@inproceedings{torabi2018behavioral,
  title={Behavioral cloning from observation},
  author={Torabi, Faraz and Warnell, Garrett and Stone, Peter},
  booktitle={International Joint Conference on Artificial Intelligence (IJCAI)},
  year={2018}
}

@inproceedings{wagenmaker2025steering,
  title={Steering Your Diffusion Policy with Latent Space Reinforcement Learning},
  author={Wagenmaker, Andrew and Nakamoto, Mitsuhiko and Zhang, Yunchu and Park, Seohong and Yagoub, Waleed and Nagabandi, Anusha and Gupta, Abhishek and Levine, Sergey},
  booktitle={Conference on Robot Learning (CoRL)},
  year={2025}
}

@inproceedings{wang2025inference,
  title={Inference-time policy steering through human interactions},
  author={Wang, Yanwei and Wang, Lirui and Du, Yilun and Sundaralingam, Balakumar and Yang, Xuning and Chao, Yu-Wei and P{\'e}rez-D’Arpino, Claudia and Fox, Dieter and Shah, Julie},
  booktitle={IEEE International Conference on Robotics and Automation (ICRA)},
  year={2025},
}

@article{wen2020fighting,
  title={Fighting copycat agents in behavioral cloning from observation histories},
  author={Wen, Chuan and Lin, Jierui and Darrell, Trevor and Jayaraman, Dinesh and Gao, Yang},
  journal={Neural Information Processing Systems (NeurIPS)},
  year={2020}
}

@inproceedings{wolczyk2024fine,
  title={Fine-tuning reinforcement learning models is secretly a forgetting mitigation problem},
  author={Wolczyk, Maciej and Cupial, Bartlomiej and Ostaszewski, Mateusz and Bortkiewicz, Michal and Zajkac, Michal and Pascanu, Razvan and Kucinski, Lukasz and Milos, Piotr},
  booktitle={International Conference on Machine Learning (ICML)},
  year={2024}
}

@inproceedings{wu2025foresight,
  title={From foresight to forethought: Vlm-in-the-loop policy steering via latent alignment},
  author={Wu, Yilin and Tian, Ran and Swamy, Gokul and Bajcsy, Andrea},
  booktitle={Robotics: Science and Systems (RSS)},
  year={2025}
}

@inproceedings{zhang2025inference,
  title={Inference-time scaling of diffusion models through classical search},
  author={Zhang, Xiangcheng and Lin, Haowei and Ye, Haotian and Zou, James and Ma, Jianzhu and Liang, Yitao and Du, Yilun},
  booktitle={International Conference on Learning Representations (ICLR)},
  year={2026}
}

@inproceedings{zhao2023learning,
  title={Learning fine-grained bimanual manipulation with low-cost hardware},
  author={Zhao, Tony Z and Kumar, Vikash and Levine, Sergey and Finn, Chelsea},
  booktitle={Robotics: Science and Systems (RSS)},
  year={2023}
}

@inproceedings{liu2023flow,
  title     = {Flow Straight and Fast: Learning to Generate and Transfer Data
               with Rectified Flow},
  author    = {Liu, Xingchao and Gong, Chengyue and Liu, Qiang},
  booktitle = {International Conference on Learning Representations (ICLR)},
  year      = {2023},
}

@inproceedings{albergo2023building,
  title     = {Building Normalizing Flows with Stochastic Interpolants},
  author    = {Albergo, Michael S. and Vanden-Eijnden, Eric},
  booktitle = {International Conference on Learning Representations (ICLR)},
  year      = {2023},
}

@inproceedings{hu2024adaflow,
  title     = {AdaFlow: Imitation Learning with Variance-Adaptive Flow-Based Policies},
  author    = {Hu, Xixi and Liu, Bo and Liu, Xingchao and Liu, Qiang},
  booktitle = {Advances in Neural Information Processing Systems 37 (NeurIPS 2024)},
  year      = {2024}
}

@inproceedings{braun2024riemannian,
  title     = {Riemannian Flow Matching Policy for Robot Motion Learning},
  author    = {Braun, Max and Jaquier, No{\'e}mie and Rozo, Leonel and Asfour, Tamim},
  booktitle = {IEEE/RSJ International Conference on Intelligent Robots and Systems (IROS)},
  year      = {2024}
}

@article{kim2021noise2score,
  title={Noise2score: tweedie’s approach to self-supervised image denoising without clean images},
  author={Kim, Kwanyoung and Ye, Jong Chul},
  journal={Advances in Neural Information Processing Systems},
  year={2021}
}

@article{efron2011tweedie,
  title={Tweedie’s formula and selection bias},
  author={Efron, Bradley},
  journal={Journal of the American Statistical Association},
  year={2011},
}
}



\end{document}